\documentclass[sigconf]{acmart}

\usepackage{tabularx}   
\usepackage{booktabs}   
\usepackage{multirow}   
\usepackage{colortbl}   

\usepackage{siunitx}    

\definecolor{mitblue}{rgb}{0.88,0.95,0.96}

\usepackage[ruled,vlined]{algorithm2e}

\usepackage{balance}

\AtBeginDocument{%
  }

\copyrightyear{2026}
\acmYear{2026}
\setcopyright{cc}
\setcctype{by}
\acmConference[MM '26]{Proceedings of the 34th ACM International Conference on Multimedia}{November 10--14, 2026}{Rio de Janeiro, Brazil}
\acmBooktitle{Proceedings of the 34th ACM International Conference on Multimedia (MM '26), November 10--14, 2026, Rio de Janeiro, Brazil}
\acmDOI{10.1145/3767308.3835447}
\acmISBN{979-8-4007-2213-4/2026/11}

\begin{document}

\title{DORS: Dynamic Attention Routing for Diffusion-based Object Removal in Dense Scenes}

\author{Haitong Tang}
\affiliation{
	\institution{School of Computer Science and Information Engineering, Hefei University of Technology}
	\city{Hefei}
	\country{China}
}
\email{httang1224@gmail.com}

\author{Haipeng Liu}
\correspondingauthor
\affiliation{
	\institution{School of Computer Science and Information Engineering,
		Hefei University of Technology}
	\city{Hefei}
	\country{China}}
\email{hpliu\_hfut@hotmail.com}

\author{Yang Wang}
\affiliation{
	\institution{School of Computer Science and Information Engineering, Hefei University of Technology}
	\city{Hefei}
	\country{China}}
\email{yangwang@hfut.edu.cn}

\begin{abstract}
	Object removal aims to eliminate target objects specified by a mask while preserving visual consistency with the surrounding regions.  
	Existing methods typically rely on contextual information from surrounding regions. 
	However, in dense scenes where the surrounding regions contain instances visually similar to the removal target, such reliance often leads to semantic interference, resulting in incomplete removal.
	This problem arises from erroneous information propagation in the attention space, where masked queries tend to align with such instances due to global similarity matching in self-attention.
	To address this challenge, we propose a \underline{\textbf{D}}iffusion-based \underline{\textbf{O}}bject \underline{\textbf{R}}emoval framework for dense \underline{\textbf{S}}cenes, dubbed \textbf{DORS},
	built upon a \textit{Dynamic Attention Routing} mechanism comprising two complementary components:  
	\textit{Instance-Filtered Attention}, which suppresses misleading semantic information from similar instances through dynamically constructed mask-guided attention constraints,
	and \textit{Context-Guided Routing}, which dynamically routes complementary scene information to maintain visual consistency.
	We further introduce DOR-Bench, a benchmark tailored for object removal in dense scenes.
	Extensive experiments demonstrate that \textbf{DORS} outperforms state-of-the-art methods, particularly in reducing incomplete removal and duplicate artifacts.
	The project page, which provides access to the source code and DOR-Bench, is publicly available at \url{https://httang1224.github.io/DORS/}. 
	
\end{abstract}



\begin{teaserfigure}
	\centering
	\includegraphics[width=\textwidth]{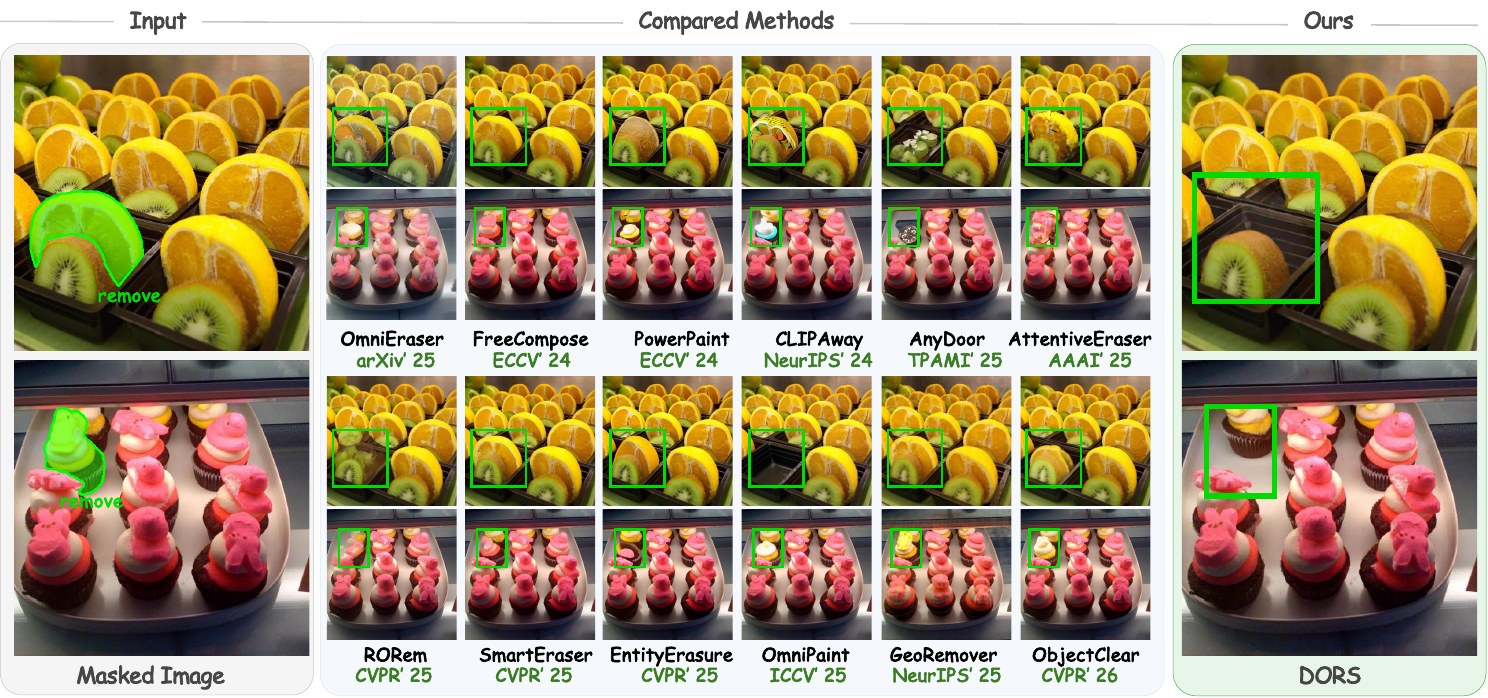}
	\caption{
		Object removal in dense scenes containing multiple similar instances remains challenging for state-of-the-art methods,
		which often fail to completely erase the target instance and leave residual artifacts, as highlighted by the green bounding boxes. In contrast, our training-free, plug-and-play DORS leverages a dynamic attention routing mechanism to fully remove the target instance while producing cleaner and more coherent results.}
	\label{fig1:teaser}
\end{teaserfigure}

\maketitle

\section{Introduction}
Given an input image and a user-provided mask, object removal aims to eliminate the specified target and reconstruct a structurally and semantically coherent background, producing a plausible target-absent scene~\cite{rorem, smarteraser, clipaway}.
The rapid proliferation of AI-generated visual content~\cite{q1111,q1112,q1113,q1114,q1115,q1116,q1117,q1118} has further intensified the demand for flexible and high-quality visual editing.
Meanwhile, recent advances in diffusion and flow-based generative models~\cite{SD, lu2025dpm, flowmatch, rectifiedflow} have substantially enhanced the generative capabilities of image inpainting and editing systems~\cite{image_inpaint_diff3, image_inpaint_diff5, image_inpaint_diff6, stableflow, zhang2025survey}, providing a strong technical foundation for object removal.

Existing object removal methods can be broadly categorized into training-based and training-free approaches.
Training-based methods
adapt pre-trained generative models to object removal using paired supervision, typically consisting of object-present images, object-removed counterparts, and corresponding target masks.
Recent studies have advanced this paradigm through large-scale synthetic data construction and augmentation~\cite{objectdrop, rorem, smarteraser, omnieraser, erasediff}, stronger conditioning signals based on textual prompts or semantic representations~\cite{powerpaint, magiceraser, clipaway, entityerasure}, and the incorporation of structural or geometric priors~\cite{georemover, asuka}.
Unified editing frameworks have also been explored to improve the general applicability and visual consistency of removal results~\cite{anydoor, omnipaint, paintbyinpaint}.
In contrast, training-free methods perform object removal by modifying the generation process of pre-trained models at inference time without additional task-specific training.
FreeCompose~\cite{freecompose} formulates object removal as a latent optimization problem and suppresses information from the masked region by discarding the corresponding key and value features in self-attention.
AttentiveEraser~\cite{attentiveeraser} instead modifies self-attention to encourage masked-region queries to aggregate information from the surrounding background.

Despite their differences, both paradigms ultimately rely on contextual information from unmasked regions to reconstruct the masked area, without explicitly distinguishing different semantic sources.
While effective in simple scenarios, this mechanism becomes unreliable in dense scenes containing multiple visually similar instances, where semantic ambiguity often leads to incomplete removal and duplicate artifacts, as highlighted by the green bounding boxes in Fig.~\ref{fig1:teaser}.
From a mechanistic perspective, this issue stems from the similarity-driven nature of self-attention: queries from masked regions tend to attend to semantically similar keys associated with unmasked instances.
As a result, semantic information from these non-target instances is erroneously propagated into the removal region, a phenomenon we term \textit{Instance Interference}.
Consequently, mask constraints or conditional guidance alone are often insufficient to reliably prevent such attention misallocation, leading to unstable and inaccurate object removal in dense scenes.

To address this fundamental limitation, we propose a \underline{\textbf{D}}iffusion-based \underline{\textbf{O}}bject \underline{\textbf{R}}emoval framework for dense \underline{\textbf{S}}cenes, named \textbf{DORS}. At its core, \textbf{DORS} introduces a dynamic attention routing mechanism that explicitly regulates information flow within the attention space. This mechanism comprises two complementary components, namely \textit{Instance-Filtered Attention} (IFA) and \textit{Context-Guided Routing} (CGR), which jointly enable fine-grained control over feature aggregation and semantic propagation. Unlike existing methods that rely on task-specific fine-tuning or implicit conditioning, \textbf{DORS} directly modulates semantic information exchange during generation, thereby effectively mitigating instance interference.
Specifically, IFA acts as a semantic shield by dynamically constructing constraints to prune misleading attention connections to semantically similar but irrelevant regions. To compensate for the resulting information vacuum, CGR dynamically redirects attention toward valid contextual cues, thereby preserving structural consistency.
Together, this complementary “shield-and-bridge” design suppresses interference while retaining meaningful background information. Furthermore, the proposed attention control is primarily applied during the early denoising stages, when the global structure is established, and is relaxed in later stages to facilitate detail refinement. By explicitly modeling and regulating information flow, \textbf{DORS} effectively mitigates semantic interference and enables accurate object removal. 

In summary, our main contributions are as follows:
{
	\leftmargini=3mm 
	\begin{itemize}
		\item We introduce a new perspective that formulates object removal as
		an information routing problem in the attention space, providing
		a principled explanation of Instance Interference in dense scenes.

		\item We propose a Dynamic Attention Routing mechanism with a complementary design of Instance-Filtered Attention and Context-Guided Routing, establishing a stage-wise control strategy that stabilizes global structure in early denoising and refines details in later steps.
		
		\item We construct a new benchmark, DOR-Bench, for object removal
		in dense scenes, and conduct extensive experiments on both
		this benchmark and public benchmarks, demonstrating that our
		method significantly outperforms existing approaches in both
		removal accuracy and visual consistency.
	\end{itemize}
}

\begin{figure*}[htbp]
	\centering
	\includegraphics[width=0.8\linewidth]{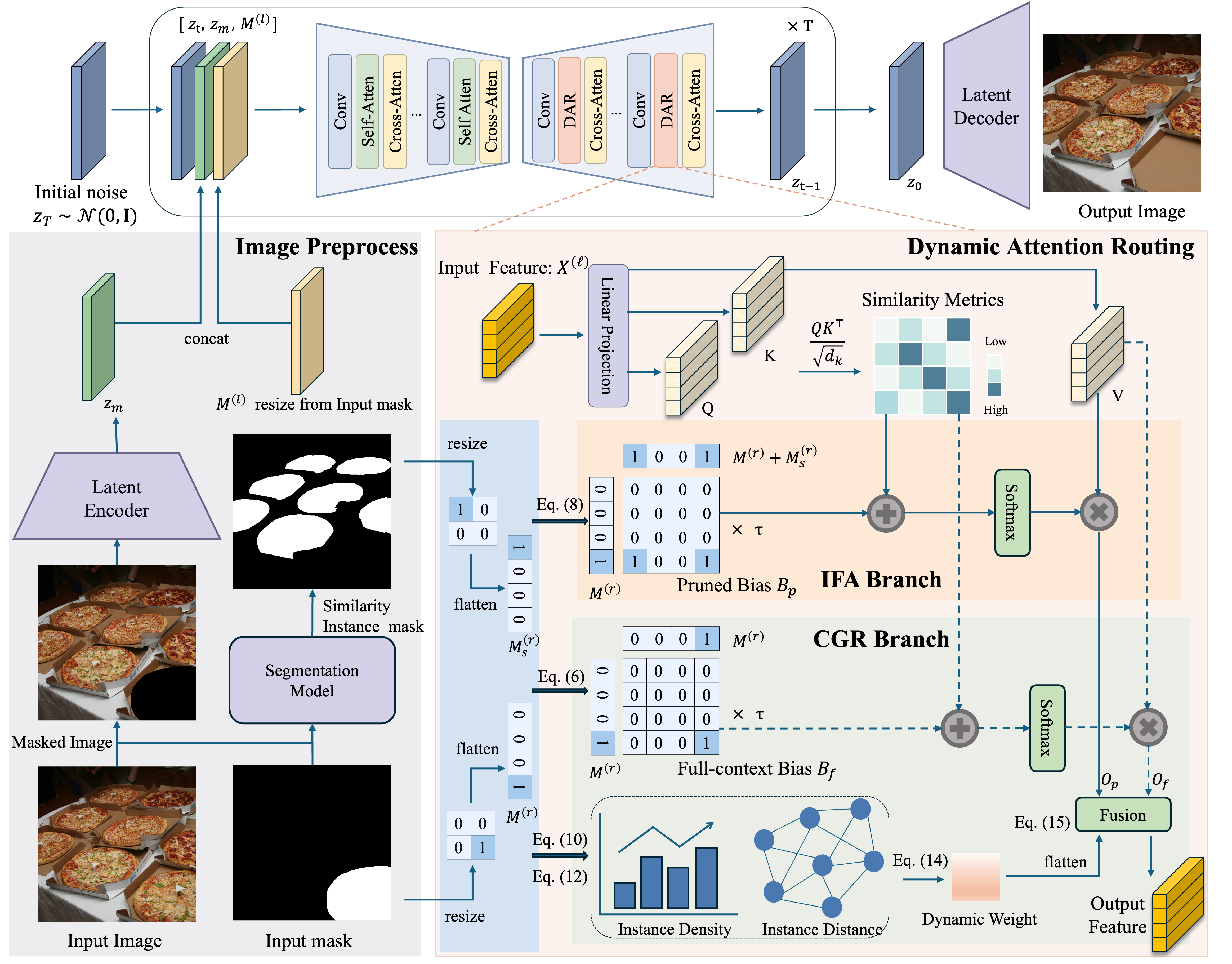}
	\caption{
		Overview of \textbf{DORS}.
		It introduces a plug-and-play Dynamic Attention Routing (DAR) mechanism into the denoising process, comprising Instance-Filtered Attention (IFA) and Context-Guided Routing (CGR).
		The former dynamically constructs mask-guided constraints to suppress similar-instance interference, while the latter dynamically routes information between the filtered and full-context pathways based on global density and local geometry. For clarity, mask resizing is simplified.}
	\label{fig2: pipe}
\end{figure*}

\section{Method} 
We propose \textbf{DORS}, a diffusion-based object removal framework for dense scenes,
built upon a Dynamic Attention Routing mechanism 
that explicitly regulates cross-region information flow. As illustrated in Fig.~\ref{fig2: pipe}, it consists of two complementary components: Instance-Filtered Attention(IFA) and Context-Guided Routing (CGR), detailed in Sec.~\ref{main:32} and ~\ref{main:33}, respectively.\textit{ Due to page limitations, the complete algorithm pipeline is provided in \textbf{ Sec. ~\ref{app:pipeline} of the Appendix}.
}

\subsection{Preliminary}
Diffusion-based image inpainting \cite{inpaint2, image_inpaint_diff2, image_inpaint_diff4} aims to fill masked regions according to visible context while preserving structural and semantic consistency. 
For efficiency, the problem is typically formulated in latent space using a variational autoencoder \cite{vae}.
Given an image $I \in \mathbb{R}^{3 \times H \times W}$ and a binary mask $M \in \{0,1\}^{1 \times H \times W}$, where $M=1$ indicates the region to be inpainted, the image is encoded as $\bar{z}_0 = E(I) \in \mathbb{R}^{C \times h \times w}$. 
During inference, the reverse process starts from $z_T \sim \mathcal{N}(0,\mathbf{I})$ and iteratively denoises the latent under conditional guidance \cite{ddpm, ddim, cfg, ddpm2}. 
To incorporate spatial context, the masked image $I_m = I \odot (1 - M)$ is encoded as $z_m = E(I_m)$, and the mask is downsampled to the latent resolution as $M^{(l)}$. 
At each timestep $t$, the noisy latent $z_t$, together with $z_m$ and $M^{(l)}$, is concatenated and fed into the noise prediction network \cite{unet, cscunet}:
\begin{equation}\label{eq:epsilon_t}
	\epsilon_t = \epsilon_\theta([z_t, z_m, M^{(l)}], t, c),
\end{equation}
where $c$ denotes optional conditioning information such as text prompts. 
The predicted noise parameterizes the reverse transition
\begin{equation}
	p_\theta(z_{t-1} \mid z_t, c) = \mathcal{N}(z_{t-1}; \mu_\theta(z_t, t, c), \sigma_t^2\mathbf{I}),
\end{equation}
which updates $z_t$ to $z_{t-1}$. 
After iterative denoising, the latent $z_0$ is decoded by $D(\cdot)$ \cite{vae} to obtain the inpainted image $D(z_0)$.

\noindent
\textbf{Self-Attention Mechanism.}
In the denoising network $\epsilon_\theta$ in Eq.~(\ref{eq:epsilon_t}), self-attention modules \cite{attention} model long-range dependencies and govern information propagation across spatial locations. 
Let $X^{(\ell)} \in \mathbb{R}^{N \times d}$ denote the feature representation at layer $\ell$. 
The features are projected into queries, keys, and values:
\begin{equation}
	Q = X^{(\ell)} W_Q, \quad K = X^{(\ell)} W_K, \quad V = X^{(\ell)} W_V,
\end{equation}
where $W_Q, W_K, W_V \in \mathbb{R}^{d \times d_k}$ are learnable projection matrices. 
The scaled dot-product similarity matrix is computed as
\begin{equation}\label{eq:similarity_matrix}
	S = \frac{QK^\top}{\sqrt{d_k}}.
\end{equation}
Applying a row-wise softmax to $S$ yields the attention weights, and the output is obtained by aggregating $V$ accordingly:
\begin{equation}\label{eq:std_qkv}
	O = \text{softmax}(S)V.
\end{equation}
Through this mechanism, each location selectively aggregates information from other positions, enabling global contextual interaction. 
Therefore, directly modifying attention weights provides a natural interface for regulating cross-region information propagation, which is particularly suitable for object removal in dense scenes.

\begin{figure}[tbp]
	\centering
	\includegraphics[width=\linewidth]{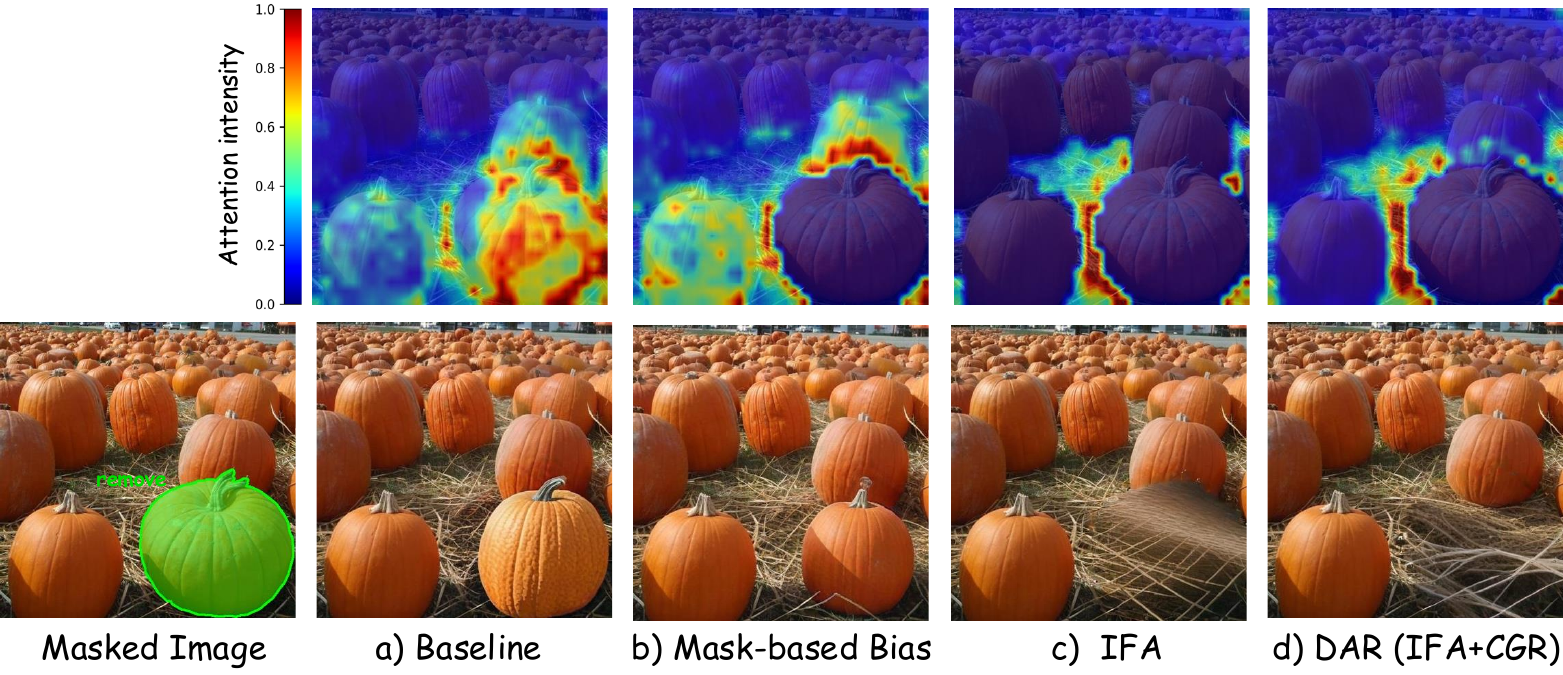}
	\caption{
		Self-attention visualization and corresponding removal results under different attention routing strategies.
		The attention maps illustrate how the masked region attends to different image regions, revealing information propagation patterns during removal.
		DAR denotes the proposed Dynamic Attention Routing mechanism.
	}
	\label{fig3: atten visul}
\end{figure}

\subsection{Instance-Filtered Attention}
\label{main:32}
When adapting diffusion-based inpainting models to object removal, a common practice is to disable textual guidance by setting $c$ to be empty in Eq.~(\ref{eq:epsilon_t}). 
However, as shown in Fig.~\ref{fig3: atten visul}(a), this strategy often fails to reliably remove the target object, since diffusion-based inpainting models are inherently trained to reconstruct missing regions rather than erase specific instances. 
Consequently, they tend to propagate target-related information into the masked region, resulting in undesired reconstruction even without textual guidance.
To mitigate this issue, prior work \cite{attentiveeraser} introduces mask-aware attention constraints that suppress interactions within the masked region. 
This can be interpreted as restricting information propagation by preventing masked queries from attending to tokens inside the masked region, thereby reducing self-reconstruction. 
Specifically, the binary mask $M$ is downsampled to match the attention resolution, yielding $M^{(r)}$. 
Let $i$ and $j$ denote the spatial positions of query and key tokens, respectively. 
The mask-based attention bias $B_f$ is defined as:
\begin{equation}\label{eq:B_f}
	B_f^{ij} =
	\begin{cases}
		\tau, & i \in M^{(r)},\ j \in M^{(r)} \\
		0, & \text{otherwise},
	\end{cases}
\end{equation}
where $\tau$ is a large negative constant that suppresses the corresponding attention weights. 
The resulting attention output is:
\begin{equation}\label{eq:O_f_}
	O_f = \text{softmax}(S + B_f)\, V.
\end{equation}
This formulation effectively reduces intra-mask information propagation and improves removal performance in simple scenarios.

However, such a formulation remains insufficient in dense scenes with multiple similar objects. 
As illustrated in Fig.~\ref{fig3: atten visul}(b), although attention within the masked region is suppressed, masked queries can still attend to instances in the surrounding regions that are visually similar to the removal target. 
These interactions introduce misleading semantic cues and often lead to duplicated structures or incomplete removal.
This observation indicates that the core failure mode of object removal in dense scenes lies not only in intra-mask information propagation, but also in erroneous aggregation of target-similar semantics from outside the mask. 
We refer to this phenomenon as \textit{instance interference}, i.e., undesired information propagation from semantically similar regions.

\begin{figure}[tbp]
	\centering
	\includegraphics[width=\linewidth]{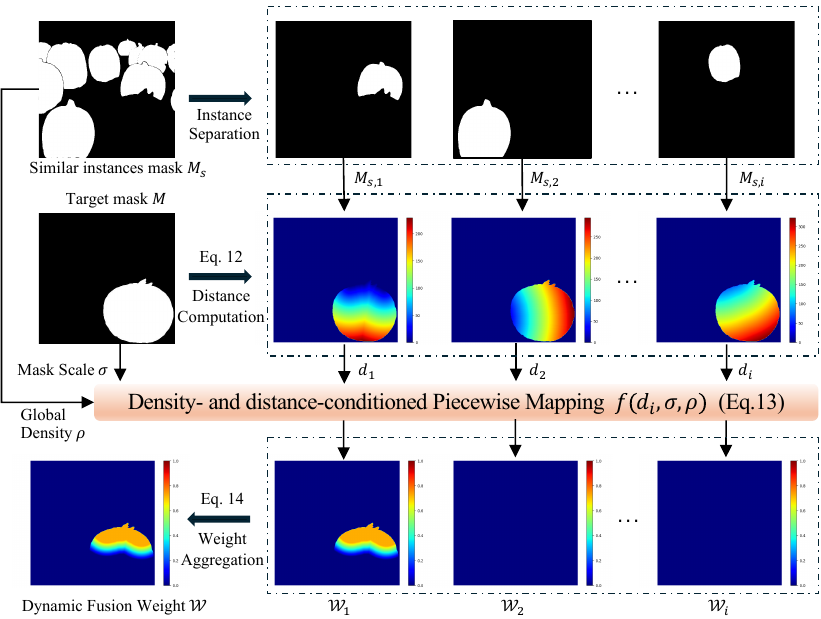}
	\caption{
		Generation of the spatially adaptive routing weight in CGR.
		The similar-instance mask is decomposed into individual instance masks, followed by distance computation to construct instance-wise filtering weights.
		A density- and geometry-aware routing function aggregates these to determine the final dynamic fusion weight.
	}
	\label{fig32:pipeline_weight}
\end{figure}
\begin{figure*}[ht]
	\centering
	\includegraphics[width=\linewidth]{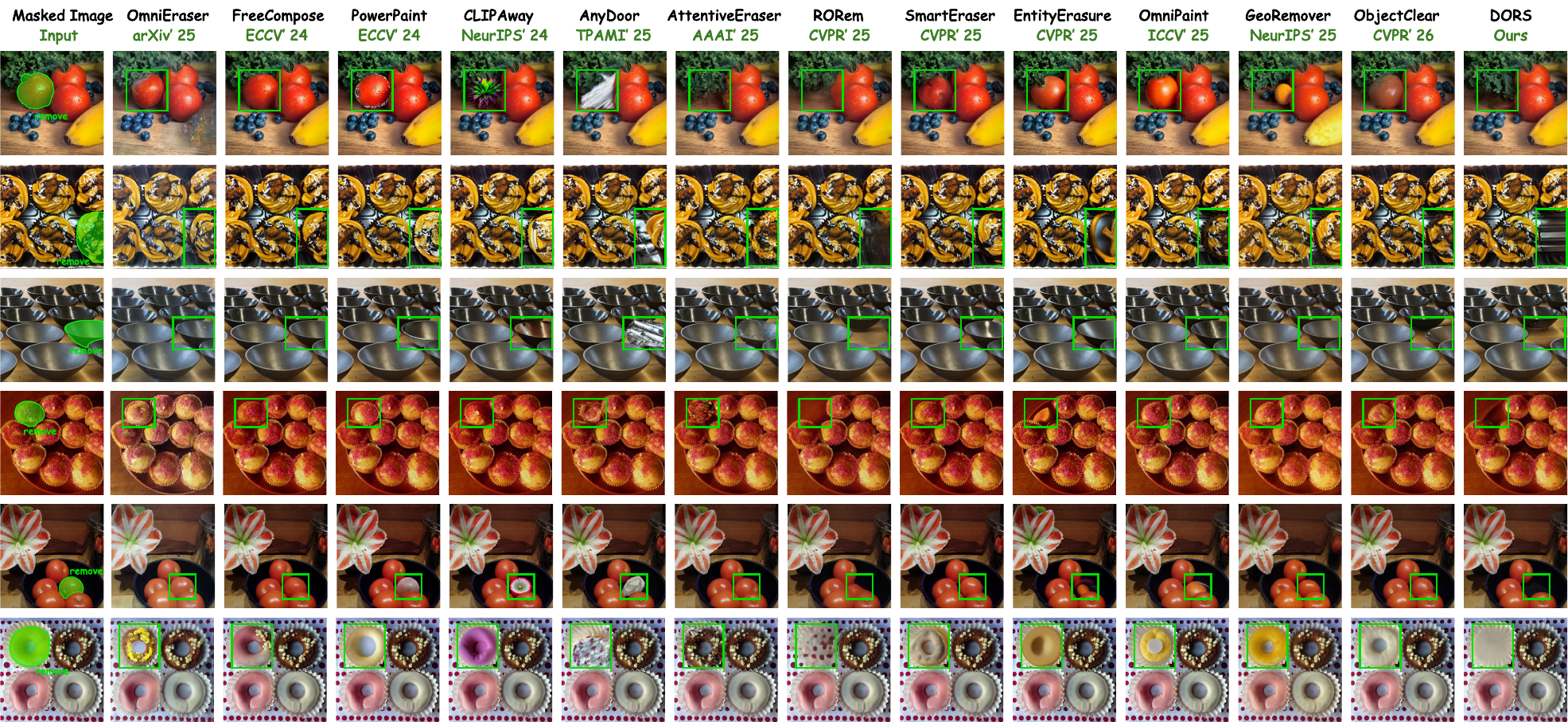}
	\caption{
		Qualitative comparison of object removal results in DOR-Bench, our method (\textbf{DORS}) achieves more accurate target removal while better preserving structural and semantic consistency, resulting in cleaner and more realistic outputs.
	}
	\label{fig4:Qualitative comparison}
\end{figure*}
To address this issue, we propose \textbf{Instance-Filtered Attention}, 
which performs semantic pathway pruning by blocking attention from the masked region to both itself and target-similar regions in the surrounding area.
Specifically, we first perform instance-level segmentation using SAM3 model \cite{sam3}, where the target removal region is used as a visual prompt to identify similar instances, 
denoted as $M_s$, and obtain its aligned representation $M_s^{(r)}$ at the attention resolution. We then introduce a similarity-aware pruned bias matrix $B_p$:
\begin{equation} \label{eq:B_p}
	B_p^{ij} =
	\begin{cases}
		\tau, & i \in M^{(r)},\ j \in \big(M^{(r)} \cup M_s^{(r)}\big) \\
		0, & \text{otherwise},
	\end{cases}
\end{equation} 

This formulation enables \textbf{dynamic, input-adaptive pruning} of attention pathways from masked queries to both intra-mask and target-similar regions, with the pruning pattern tailored to instance configuration of each input image, thereby effectively suppressing undesired semantic propagation.
The resulting attention output is:
\begin{equation} \label{eq:O_p}
	O_p = \text{softmax}(S + B_p)\, V.
\end{equation}

By removing both intra-mask and inter-instance attention pathways, masked queries are constrained to aggregate information only from non-similar regions in the surrounding context.  
This effectively eliminates misleading semantic references and mitigates instance interference in dense scenes. 
As shown in Fig.~\ref{fig3: atten visul}(c), this leads to consistent object removal without regenerating target-related structures. 
Importantly, the proposed suppression is applied only to masked queries, thereby preserving valid information propagation within the unmasked background.

\subsection{Context-Guided Routing}
\label{main:33}

While Instance-Filtered Attention suppresses misleading semantic propagation, it may also limit access to useful structural context. 
As shown in Fig.~\ref{fig3: atten visul}(c), 
although the target is removed, the area above the mask exhibits noticeable artifacts,
indicating that overly aggressive filtering may hinder access to necessary contextual information for coherent reconstruction. 
This suggests that pruning attention to similar instances, while effective at eliminating undesired semantics, may also discard informative structural context.
Therefore, the key challenge lies not 
in completely suppressing such information, but in selectively and \textbf{dynamically regulating} its propagation according to the spatial relationships between the target region and surrounding similar instances.

To address this, we formulate object removal as an \textit{information routing} problem in the attention space.
The goal is to dynamically control information propagation at each spatial location, balancing structural cues and misleading semantics.
Specifically, we consider two complementary pathways:
a full-context branch ($O_f$ in Eq.~\eqref{eq:O_f_}), derived from the mask-based bias (Fig.~\ref{fig3: atten visul}(b)), which suppresses intra-mask propagation while preserving access to the unmasked context,
and an instance-filtered branch ($O_p$ in Eq.~\eqref{eq:O_p}), which blocks interactions with target-similar regions (Fig.~\ref{fig3: atten visul}(c)).
We then introduce a dynamic, spatially adaptive routing mechanism that fuses the two branches according to the global distribution and local geometric relationships of target-similar instances.

Our design is guided by two complementary considerations.
First, from a global perspective,
a greater concentration of target-similar instances leads to stronger semantic interference, necessitating more aggressive suppression of misleading information.
Second, from a local geometric perspective,
the distance to target-similar instances determines the trade-off between structural continuity and semantic interference: spatial locations near similar instances require greater access to contextual information to preserve local coherence, whereas distant locations favor stronger suppression to prevent erroneous semantic propagation.

Based on the first consideration, we define a global density measure to characterize the overall concentration of target-similar instances in the current scene:
\begin{equation}\label{eq:rho}
	\rho =
	\frac{\text{area}(M_s)}
	{\text{area}(\text{unmasked region})},
\end{equation}
where a larger $\rho$ indicates a higher concentration of target-similar instances and stronger semantic interference.
Accordingly, the maximum filtering strength is defined as:
\begin{equation}\label{eq:w_max}
	w_{\max}=\alpha\rho,
\end{equation}
where $\alpha\in[0,1]$ controls influence of global instance distribution.

Based on the second principle, we further model local spatial relationships in a geometry-aware manner.
For each pixel $x$ in the masked region, we compute its distance to nearby similar instances:
\begin{equation}\label{eq:d_i}
	d_i(x)=\text{dist}(x,M_{s,i}),
\end{equation}
where $M_{s,i}$ denotes the $i$-th instance in $M_s$.

To combine these two principles, we assign lower filtering weights to proximal regions to preserve structural continuity, and progressively increase them for distant regions where contextual information is more likely to introduce semantic interference.
Let $\sigma$ denote the diameter of the target mask.
We introduce two hyperparameters $\beta_1$ and $\beta_2$ ($0<\beta_1<\beta_2<1$) to control the transition range, where $\beta_1$ defines the region requiring full-context information and $\beta_2$ determines the suppression boundary.
Accordingly, we define a adaptive filtering weight that jointly accounts for global instance density and local distance to target-similar instances as follows:

\begin{equation}\label{eq:D_i}
	w_i(x)=
	\begin{cases}
		0, & d_i(x)\leq\beta_1\sigma,\\[4pt]
		w_{\max}
		\frac{d_i(x)-\beta_1\sigma}
		{(\beta_2-\beta_1)\sigma},
		& \beta_1\sigma<d_i(x)<\beta_2\sigma,\\[6pt]
		w_{\max}, & d_i(x)\geq\beta_2\sigma .
	\end{cases}
\end{equation}

To aggregate multiple instances, we take the maximum response:
\begin{equation}\label{eq:w_p}
	w(x)=\max_i w_i(x).
\end{equation}

\begin{table*}[t]
	\centering
	\caption{Quantitative Comparisons with State-of-the-Arts on the proposed DOR-Bench. The best value for each metric is highlighted in \textcolor{red}{\textbf{red bold}}, and the second best is shown as \textcolor{blue}{\textbf{blue bold}}.
	}
	\label{tab1:dor}
	\small
	\setlength{\tabcolsep}{3.3pt}       

	\begin{tabularx}{\textwidth}{l c | *{3}{>{\centering\arraybackslash}X} | *{2}{>{\centering\arraybackslash}X} | *{3}{>{\centering\arraybackslash}X}}
		\toprule
		\multirow{2}{*}{Method} & \multirow{2}{*}{Type} 
		& \multicolumn{3}{c|}{Removal Accuracy} 
		& \multicolumn{2}{c|}{Visual Quality} 
		& \multicolumn{3}{c}{Background Fidelity} \\
		\cmidrule(lr){3-5} \cmidrule(lr){6-7} \cmidrule(lr){8-10}
		& & ReMOVE $\uparrow$ & MSN $\downarrow$ & MARS $\downarrow$ 
		& FID $\downarrow$ & AS $\uparrow$ 
		& PSNR $\uparrow$ & LPIPS $\downarrow$ & MSE $\downarrow$ \\
		\midrule
		
		PowerPaint (ECCV'24) \cite{powerpaint} & Training-Based & 73.08 & 55.00 & 43.14 & 49.39 & 6.05 & 22.05 & 87.79 & 63.56 \\
		CLIPAway (NeurIPS'24) \cite{clipaway} & Training-Based & 68.93 & 68.25 & 49.14 & 31.05 & 6.31 & 21.33 & 49.53 & 10.27 \\
		ObjectRemovalFluxFill \cite{object-removal-lora}& Training-Based & 71.73 & 49.00 & 42.84 & 45.71 & 6.22 & 20.56 & 80.97 & 11.62 \\
		OmniEraser (arXiv'25) \cite{omnieraser}& Training-Based & 71.16 & 40.75 & 32.01 & 93.14 & 5.63 & 13.76 & 175.91 & 48.10 \\
		AnyDoor (TPAMI'25) \cite{anydoor}& Training-Based & 71.11 & 31.25 & 10.35 & 16.96 & 6.28 & 25.14 & 18.14 & 5.89 \\
		SmartEraser (CVPR'25) \cite{smarteraser}& Training-Based & 71.97 & 57.50 & 46.44 & 41.06 & 6.15 & 21.11 & 98.80 & 24.15 \\
		EntityErasure (CVPR'25) \cite{entityerasure}& Training-Based & 71.58 & 44.50 & 35.45 & 27.94 & 6.37 & 21.46 & 49.87 & 9.77 \\
		RORem (CVPR'25) \cite{rorem}& Training-Based & \textcolor{blue}{\textbf{73.25}} &\textcolor{blue}{\textbf{6.50}} &\textcolor{blue}{\textbf{3.60}}  & 20.71 & 6.26 & 22.86 & 37.51 & 6.07 \\
		OmniPaint (ICCV'25) \cite{omnipaint}& Training-Based & 72.48 & 27.75 & 18.84 & 19.09 & \textcolor{red}{\textbf{6.47}} & 24.36 & 24.58 & 4.51 \\
		GeoRemover (NeurIPS'25) \cite{georemover} & Training-Based & 72.63 & 44.75 & 37.10 & 38.79 & 6.08 & 19.11 & 86.04 & 15.80 \\
		ObjectClear (CVPR'26) \cite{objectclear} & Training-Based & 72.91 & 28.75 & 21.34 &  \textcolor{red}{\textbf{11.67}} & \textcolor{blue}{\textbf{6.46}} &	\textcolor{blue}{\textbf{29.83}} &	\textcolor{blue}{\textbf{7.43}}&	\textcolor{blue}{\textbf{1.90}}  \\
		\midrule  
		FLUX.1-Fill-dev \cite{flux2024}& Training-Free & 69.10 & 64.75 & 55.74 & 25.00 & 6.43 & 23.60 & 40.75 & 5.09 \\
		FreeCompose (ECCV'24) \cite{freecompose} & Training-Free & 70.20 & 57.75 & 50.43 & 23.05 & 6.39 & 21.26 & 45.22 & 10.33 \\
		AttentiveEraser (AAAI'25) \cite{attentiveeraser}& Training-Free & 67.45 & 14.00 & 6.31 & 33.68 & 6.31 & 20.36 & 54.85 & 11.74 \\
		
		\cellcolor{mitblue}\textbf{DORS (Ours) }&  \cellcolor{mitblue}Training-Free & \cellcolor{mitblue} \textcolor{red}{\textbf{75.53}} & \cellcolor{mitblue}\textcolor{red}{\textbf{1.25}} & \cellcolor{mitblue}\textcolor{red}{\textbf{0.70}}  & \cellcolor{mitblue} \textcolor{blue}{\textbf{12.93}} & \cellcolor{mitblue}\textcolor{blue}{\textbf{6.46}}&  \cellcolor{mitblue}\textcolor{red}{\textbf{35.06}} & \cellcolor{mitblue}\textcolor{red}{\textbf{3.14}}  & \cellcolor{mitblue}\textcolor{red}{\textbf{0.53}}  \\
		\bottomrule
	\end{tabularx}
\end{table*}

This design assumes that the strongest filtering requirement among all target-similar instances determines the local routing weight.
Therefore, maximum aggregation preserves the dominant semantic interference signal while avoiding dilution from irrelevant instances.
The overall computation of $w(x)$ is illustrated in Fig.~\ref{fig32:pipeline_weight}.
This spatially adaptive weight controls the contribution of the filtered pathway at location $x$.
The final attention output is computed as:

\begin{equation}\label{eq:O_final}
	O=w(x)\,O_p+(1-w(x))\,O_f,
\end{equation}
resulting in a distribution- and geometry-aware routing mechanism.
Specifically, the full-context pathway is favored in proximal regions to maintain structural continuity, whereas the filtered pathway dominates distant regions to suppress misleading semantic propagation.
As shown in Fig.~\ref{fig3: atten visul}(d), this effectively mitigates artifacts while maintaining global coherence.

\noindent
\textbf{Stage-wise Application.}
In diffusion models, early denoising steps primarily establish global structure, while later steps refine local details. 
Based on this observation, we apply the proposed routing mechanism only in the early stages of the reverse process (e.g., the first half of timesteps), and revert to the standard predictor in later stages for detail refinement. 
This stage-wise design effectively balances structural consistency and fine-grained detail generation, as validated in our ablation study. 
To further preserve background fidelity, we incorporate latent blending as a complementary safeguard, ensuring that unmasked regions remain unchanged.

\section{Experiments}
\subsection{Implementation Details}
\textbf{Benchmarks.}
Since existing benchmarks lack systematic evaluation for object removal in dense scenes, we construct a new benchmark, termed \textit{Dense Object Removal Benchmark (DOR-Bench)}, consisting of 400 image--mask pairs covering diverse scenes with multiple similar instances. 
All images are manually curated, and masks are carefully annotated and refined to ensure reliable evaluation. 
In addition, following previous works \cite{omnipaint, objectclear, omnieraser}, we evaluate on the RORD benchmark \cite{rord} with 500 image--mask pairs for fair comparison and to assess generalization. 
\textit{Due to page limitations, see the detailed construction of DOR-Bench in
	\textbf{ Sec. ~\ref{app:dor} of the Appendix}	
	.}

\noindent
\textbf{Baselines.}
We compare our method with a broad range of state-of-the-art object removal methods, including both training-based and training-free approaches. 
The former include diffusion-based methods such as PowerPaint \cite{powerpaint}, CLIPAway \cite{clipaway}, AnyDoor \cite{anydoor}, SmartEraser \cite{smarteraser}, EntityErasure \cite{entityerasure}, RORem \cite{rorem}, and ObjectClear \cite{objectclear}, as well as recent flow-based models such as ObjectRemovalFluxFill \cite{object-removal-lora}, OmniEraser \cite{omnieraser}, OmniPaint \cite{omnipaint}, and GeoRemover \cite{georemover}. 
The latter include FreeCompose~\cite{freecompose} and the AttentiveEraser~\cite{attentiveeraser}, which operate on pre-trained models at inference. 
We also include FLUX.1-Fill-dev~\cite{flux2024} as a strong inpainting baseline.

\noindent
\textbf{Evaluation Metrics.}
We evaluate performance from three perspectives:
(1) \textit{Removal Accuracy}, including ReMOVE \cite{remove}, MSN \cite{entityerasure}, and MARS \cite{entityerasure}, where ReMOVE measures feature consistency between the removed region and the background, MSN counts newly generated spurious objects, and MARS evaluates the area ratio of artifacts within the masked region;
(2) \textit{Visual Quality}, including FID \cite{fid} and Aesthetic Score (AS) \cite{as}, where FID measures the distribution discrepancy between generated and real images, while AS reflects perceptual visual quality;
(3) \textit{Background Fidelity}, including PSNR \cite{psnr}, LPIPS \cite{lpips}, and MSE \cite{mse}, which measure background consistency from both pixel-level and perceptual perspectives.

\begin{table*}[t]
	\centering
	\caption{Quantitative Comparisons with State-of-the-Arts on the public RORD Benchmark \cite{rord}. The best value for each metric is highlighted in \textcolor{red}{\textbf{red bold}}, and the second best is shown as \textcolor{blue}{\textbf{blue bold}}. }
	\label{tab2:rord}
	\small
	\setlength{\tabcolsep}{3.3pt}       

	\begin{tabularx}{\textwidth}{l c | *{3}{>{\centering\arraybackslash}X} | *{2}{>{\centering\arraybackslash}X} | *{3}{>{\centering\arraybackslash}X}}
		\toprule
		\multirow{2}{*}{Method} & \multirow{2}{*}{Type} 
		& \multicolumn{3}{c|}{Removal Accuracy} 
		& \multicolumn{2}{c|}{Visual Quality} 
		& \multicolumn{3}{c}{Background Fidelity} \\
		\cmidrule(lr){3-5} \cmidrule(lr){6-7} \cmidrule(lr){8-10}
		& & ReMOVE $\uparrow$ & MSN $\downarrow$ & MARS $\downarrow$ 
		& FID $\downarrow$ & AS $\uparrow$ 
		& PSNR $\uparrow$ & LPIPS $\downarrow$ & MSE $\downarrow$ \\
		\midrule
		
		PowerPaint (ECCV'24) \cite{powerpaint}& Training-Based & 89.49 & 62.80 & 9.63 & 46.50 & 4.35& \textcolor{blue}{\textbf{23.44}} & 56.63 & 21.26 \\
		CLIPAway (NeurIPS'24) \cite{clipaway} & Training-Based & 85.78 & 148.40 & 14.67 & 51.61 & 4.66 & 21.99 & 47.88 & 9.79 \\
		ObjectRemovalFluxFill \cite{object-removal-lora} & Training-Based & 90.73 & 31.60 & 6.30 & 48.11 & 4.27 & 20.76 & 72.98 & 9.53 \\
		OmniEraser (arXiv'25) \cite{omnieraser}& Training-Based & 92.24 & 38.40 & 1.39 & 50.92 & 4.34 & 13.68 & 128.63 & 49.51 \\
		AnyDoor (TPAMI'25) \cite{anydoor}& Training-Based & 79.23 & 45.20 & 2.10 & 112.76 & 4.34 & 15.20 & 102.56 & 44.78 \\
		SmartEraser (CVPR'25) \cite{smarteraser}& Training-Based & 92.64 & 24.80 & 1.01 & 25.87 & 4.58 & 19.20 & 59.89 & 16.48 \\
		EntityErasure (CVPR'25) \cite{entityerasure}& Training-Based & 89.04 & 38.20 & 8.26 & 42.26 & 4.64 & 22.21 & 50.12 & 9.3 \\
		RORem (CVPR'25) \cite{rorem}& Training-Based & 92.37 & \textcolor{blue}{\textbf{15.40}}  & 1.26 & 34.27 & 4.44 & 20.27 & 75.54 & 11.71 \\
		OmniPaint (ICCV'25) \cite{omnipaint} & Training-Based & 91.46 & 24.20 & 2.43 & 21.57 & 4.66 & 20.96 & 33.04 & 11.50 \\
		GeoRemover (NeurIPS'25) \cite{georemover}& Training-Based & 87.93 & 19.80 & 0.77 & 247.31 & 3.30 & 5.02 & 505.25 & 332.75 \\
		ObjectClear (CVPR'26) \cite{objectclear}& Training-Based &\textcolor{blue}{\textbf{92.66}} & 26.80 & \textcolor{blue}{\textbf{0.74}} & \textcolor{red}{\textbf{18.05}} & 4.63 & 23.26 &  \textcolor{blue}{\textbf{19.49}}  & 8.21 \\
		\midrule  
		FLUX.1-Fill-dev \cite{flux2024}& Training-Free & 73.47 & 123.00 & 39.91 & 106.51 & \textcolor{red}{\textbf{4.84}} & 23.42 & 37.35 & \textcolor{blue}{\textbf{5.60}} \\
		FreeCompose (ECCV'24) \cite{freecompose} & Training-Free & 87.15 & 97.40 & 5.11 & 52.65 & 4.48 & 19.64 & 57.48 & 13.50 \\
		AttentiveEraser (AAAI'25) \cite{attentiveeraser} & Training-Free & 82.29 & 50.80 & 1.18 & 101.89 & 4.68 & 20.42 & 62.54 & 12.49 \\
		\cellcolor{mitblue}\textbf{DORS (Ours)} & \cellcolor{mitblue} Training-Free & \cellcolor{mitblue}\textcolor{red}{\textbf{92.69}} &  \cellcolor{mitblue}\textcolor{red}{\textbf{9.80}} & \cellcolor{mitblue} \textcolor{red}{\textbf{0.50}} &  \cellcolor{mitblue}\textcolor{blue}{\textbf{21.13}} & \cellcolor{mitblue} \textcolor{blue}{\textbf{4.70}} &  \cellcolor{mitblue}\textcolor{red}{\textbf{31.58}} &  \cellcolor{mitblue}\textcolor{red}{\textbf{5.02}} & \cellcolor{mitblue} \textcolor{red}{\textbf{0.99}} \\
		\bottomrule
	\end{tabularx}

\end{table*}

\noindent
\textbf{Implementation Details.}
Our method is built upon the SDXL inpainting model~\cite{sdxl} and requires no additional training. 
The proposed dynamic attention routing modules are applied to the decoder of the denoising network by modifying self-attention modules. 
We use 20 denoising steps with empty text conditioning by default. 
Similar instance masks are obtained using SAM3~\cite{sam3}. 
All experiments are conducted on a single NVIDIA A100 GPU.
For fair comparison, all baselines are reproduced using their official implementations and recommended configurations.

\begin{table}[t]
	\centering
	\caption{Plug-and-play evaluation across different backbones (SD1.5, SD2.0, SDXL). 
		We report MSN and MARS before and after applying our method. 
		Performance gains are shown in parentheses, where negative values indicate improvement.}
	\label{tab8:plug_and_play}
	\begin{tabular}{c c c c}
		\toprule
		Backbone & Setting & MSN $\downarrow$ & MARS $\downarrow$ \\
		\midrule
		
		\multirow{2}{*}{SD1.5} 
		& Baseline  & 58.00 & 41.49 \\
		&  \cellcolor{mitblue}Baseline +\textbf{ Ours} 
		&  \cellcolor{mitblue}16.25 (\textbf{-41.75}) 
		&  \cellcolor{mitblue}5.21 (\textbf{-36.28}) \\
		
		\midrule
		
		\multirow{2}{*}{SD2.0} 
		& Baseline & 64.25 & 44.32 \\
		&  \cellcolor{mitblue}Baseline + \textbf{Ours} 
		&  \cellcolor{mitblue}18.61 (\textbf{-45.64}) 
		&  \cellcolor{mitblue}6.75 (\textbf{-37.57}) \\
		
		\midrule
		
		\multirow{2}{*}{SDXL}  
		& Baseline & 47.75 & 33.23 \\
		&  \cellcolor{mitblue}Baseline + \textbf{Ours} 
		&  \cellcolor{mitblue}1.25 (\textbf{-46.50})  
		&  \cellcolor{mitblue}0.70 (\textbf{-32.53}) \\
		
		\bottomrule
	\end{tabular}
\end{table}

\subsection{Comparison with State-of-the-Arts}
\textbf{Quantitative Analysis.}
As shown in Tab.~\ref{tab1:dor}, our method achieves the best performance on most metrics of DOR-Bench. Notably, in terms of removal accuracy, our method achieves a ReMOVE score of 75.53, while significantly reducing MSN and MARS to 1.25 and 0.70, respectively. 
Compared to the second-best results (73.25 for ReMOVE, 6.50 for MSN, and 3.60 for MARS), our method demonstrates clear advantages across all removal-related metrics, indicating more accurate and complete object removal with fewer spurious artifacts. 
For background fidelity, our method also outperforms the best competing method ObjectClear \cite{objectclear}, improving PSNR from 29.83 to 35.06, while reducing LPIPS from 7.43 to 3.14 and MSE from 1.90 to 0.53. 
This indicates more accurate preservation of structural and textural consistency in the original background.
For visual quality, our method achieves competitive results on FID and AS, while maintaining a strong balance between removal effectiveness and visual realism in a training-free setting. 

We further evaluate generalization on the RORD benchmark \cite{rord}, as shown in Tab.~\ref{tab2:rord}. 
Our method maintains strong performance across datasets, achieving the best removal accuracy and background fidelity while preserving visual quality. 
This demonstrates that our approach generalizes well to different data distributions, benefiting from the proposed dynamic attention mechanism.

\noindent
\textbf{Qualitative Analysis.}
Fig.~\ref{fig4:Qualitative comparison} presents qualitative comparisons on DOR-Bench. 
In dense scenes, existing methods exhibit typical failure modes. 
\emph{Incomplete removal:}
Some methods fail to entirely remove the target, leaving residual structures. 
For example, OmniPaint \cite{omnipaint} (row 3) fails to fully erase the bowl structures, while ObjectClear \cite{objectclear} (row 5) retains the tomato shape and color. 
\emph{Hallucinated artifacts:}
New object-like structures are generated in the masked region. 
For instance, EntityErasure \cite{entityerasure} (row 2) produces unnatural artifacts, and AttentiveEraser \cite{attentiveeraser} (row 6) introduces a blurred donut-like artifact. 
\emph{Over-smoothing:}
Excessive smoothing leads to loss of texture details, as observed in OmniEraser \cite{omnieraser} (row 1) and RORem \cite{rorem} (row 2). 
\emph{Background inconsistency:}
Some methods unintentionally modify unmasked regions. 
GeoRemover \cite{georemover} (row 1) alters object colors, and OmniEraser \cite{omnieraser} (row 4) changes object appearance, consistent with weaker background fidelity.
These issues stem from \emph{instance interference}, where masked regions attend to similar instances and introduce misleading information, leading to inferior removal accuracy (lower ReMOVE and higher MSN/MARS) and degraded background fidelity (lower PSNR and higher LPIPS).
In contrast, our method achieves more complete removal with fewer artifacts while preserving structural and textural consistency across diverse scenes. 
This improvement is enabled by explicitly controlling attention propagation, effectively suppressing interference from similar instances. 
\textit{Due to page limitations, more results on both benchmarks are provided in 
	\textbf{ Sec. ~\ref{app:more_results} of the Appendix}.}

\noindent
\textbf{Plug-and-Play Evaluation.}
We evaluate the plug-and-play capability of our method across different backbones (SD1.5~\cite{SD}, SD2.0 \cite{SD}, and SDXL \cite{sdxl}), as shown in Tab.~\ref{tab8:plug_and_play}.
Our method consistently improves performance across all three backbones, substantially reducing both MSN and MARS.
These results demonstrate the strong generalizability of our method across different backbones.

\noindent
\textbf{User Study.}
Following prior works \cite{smarteraser, layer_decomposition, omnieraser}, we conduct a user study with human participants and a Vision Language Model (VLM)~\cite{qwen3.5} as evaluators.
Given outputs of our method and four competing baselines, evaluators select the best result based on three criteria: removal quality, background preservation, and overall image quality.
As shown in Tab.~\ref{tab9:user study}, our method is consistently preferred by both human annotators and the VLM across all criteria, demonstrating superior removal performance and strong agreement between human and automated evaluations. 
\textit{Due to page limitations, detailed user study design, evalution criteria and assessment protocol are provided in 
	\textbf{ Sec. ~\ref{app:user_study} of the Appendix}.}

\noindent
\textbf{Efficiency Analysis.}
We evaluate efficiency in terms of inference time and GPU memory.
Our method achieves competitive or lower cost than training-based approaches without additional training, while remaining comparable to training-free methods.
The reported efficiency of the full \textbf{DORS} includes the overhead of the integrated SAM3 segmentation module.
It requires 4.85s and 10.50GB VRAM per image, compared to 3.43s for AttentiveEraser~\cite{attentiveeraser} and 24.26s for FreeCompose~\cite{freecompose}.
Overall, our method achieves competitive efficiency.
\textit{Due to page limitations, detailed efficiency comparisons with existing removal methods are provided in \textbf{Sec.~\ref{app:efficiency} of the Appendix}.}

\begin{figure}[t]
	\includegraphics[width=0.85\linewidth]{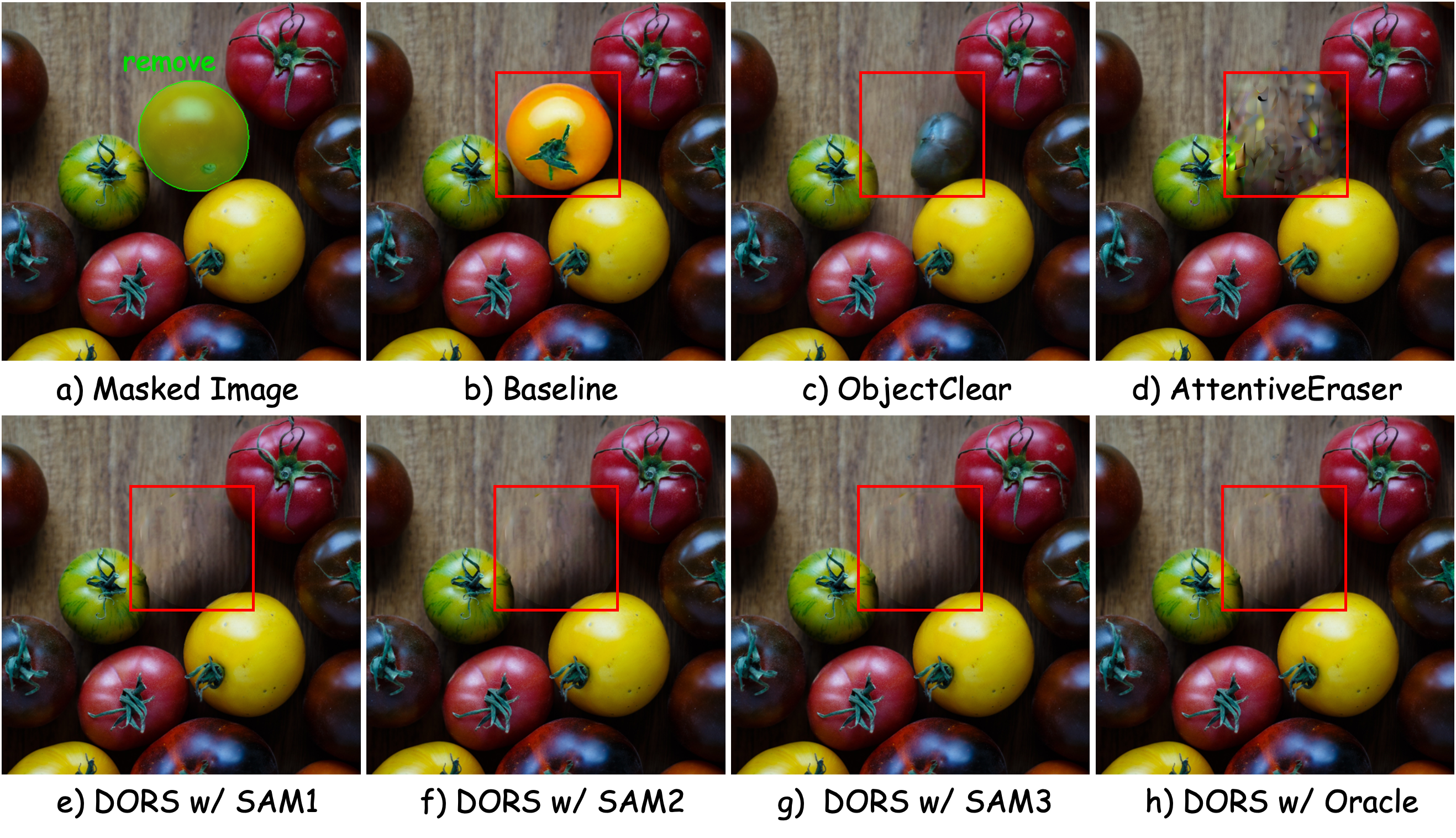}
	\caption{
		Qualitative comparison of DORS with different segmentation models in challenging dense scenes.
	}
	\label{fig:mask_source}
\end{figure}

\subsection{Ablation Studies}
\textbf{Component Ablation.}
As shown in Tab.~\ref{tab3:ablation_components}, the baseline without attention control performs poorly, with an MSN of 47.75 and a MARS of 33.23. Adding the standard full-context branch (FCB) provides only limited improvement, whereas Instance-Filtered Attention (IFA) substantially reduces MSN and MARS to 11.75 and 5.43, respectively, with moderate computational overhead. Building upon IFA, Context-Guided Routing (CGR) further improves the scores to 1.25 and 0.70 with only a marginal increase in inference time and no additional memory. These results confirm that IFA suppresses misleading information from similar instances, while CGR selectively restores useful contextual cues for coherent reconstruction, as further illustrated in Fig.~\ref{fig3: atten visul}.
\textit{Due to page limitations, additional qualitative ablation results are provided in \textbf{Sec.~\ref{app:ablation} of the Appendix}, while detailed design ablations and parameter sensitivity analyses are presented in \textbf{Sec.~\ref{app:designablation} and~\ref{app:parameter} of the Appendix}.}

\begin{table}[t]

	\centering
	\caption{\textbf{User study results based on multi-method preference evaluation. We report the percentage of selections by human participants and a vision-language model (VLM).} }
	\label{tab9:user study}
	\small
	\renewcommand{\arraystretch}{1.2}
	\setlength{\tabcolsep}{6pt}
	\begin{tabular}{l c c}
		\toprule
		Method & Human (\%) ↑ & VLM (\%) ↑ \\
		\midrule
		
		RORem (CVPR’25)~\cite{rorem}           & 8.34 & 8.45 \\
		OmniPaint (ICCV’25)~\cite{omnipaint}      & 9.01 & 7.11 \\
		GeoRemover (NeurIPS'25)~\cite{georemover}   & 2.11 & 1.48 \\
		ObjectClear (CVPR’26)~\cite{objectclear}    & 2.45 & 2.21 \\
		
		\cellcolor{mitblue}\textbf{DORS (Ours)} &  \cellcolor{mitblue}\textbf{78.09} &  \cellcolor{mitblue}\textbf{80.75 }\\
		\bottomrule
	\end{tabular}
\end{table}

\noindent
\textbf{Effect of Segmentation Models.}
We evaluate \textbf{DORS} under varying segmentation quality on a challenging dense-scene subset, using similar-instance masks in the unmasked regions generated by SAM1~\cite{sam1}, SAM2~\cite{sam2}, and SAM3~\cite{sam3}, together with manually annotated Oracle masks. As shown in Fig.~\ref{fig:mask_source}, \textbf{DORS} successfully removes the target across all three segmentation models, whereas existing methods often fail to achieve complete removal.
The quantitative results in the Appendix further show that the removal performance of \textbf{DORS} improves as the segmentation masks become more accurate. Notably, even with the relatively weaker SAM1~\cite{sam1}, \textbf{DORS} still achieves strong removal performance and outperforms ObjectClear and AttentiveEraser, demonstrating its robustness to imperfect mask inputs. \textit{Due to page limitations, detailed quantitative comparisons and analyses are provided
	rein \textbf{Sec. ~\ref{app:segmentation} of the Appendix}.}

\begin{table}[t]
	\centering
	\caption{
		Ablation of key components. 
		FCB denotes standard full-context attention, while IFA and CGR are our Instance-Filtered Attention and Context-Guided Routing.
	}
	\label{tab3:ablation_components}
	
	\small
	\renewcommand{\arraystretch}{1.11}
	\setlength{\tabcolsep}{4pt}
	\begin{tabularx}{\columnwidth}{
			>{\centering\arraybackslash}p{0.5cm}
			>{\centering\arraybackslash}p{0.6cm}
			>{\centering\arraybackslash}p{0.6cm}
			>{\centering\arraybackslash}p{0.6cm}
			| *{2}{>{\centering\arraybackslash}X}
			| *{2}{>{\centering\arraybackslash}X}
		}
		\toprule
		\multicolumn{4}{c|}{Components} 
		& \multicolumn{2}{c|}{Accuracy} 
		& \multicolumn{2}{c}{Efficiency} \\
		
		\cmidrule(lr){1-4}\cmidrule(lr){5-6}\cmidrule(lr){7-8}
		
		& FCB & IFA & CGR 
		& MSN $\downarrow$ & MARS $\downarrow$ 
		& Time $\downarrow$ & GPUM $\downarrow$ \\
		
		\midrule
		
		(a) & \ding{55} & \ding{55} & \ding{55} 
		& 47.75 & 33.23 
		& 1.51 & 7.17 \\
		
		(b) & \ding{51} & \ding{55} & \ding{55} 
		& 33.50 & 24.83 
		& 3.10 & 7.17 \\
		
		(c) & \ding{55} & \ding{51} & \ding{55} 
		& 11.75 & 5.43  
		& 4.67 & 10.50 \\
		\cellcolor{mitblue}(d) &  \cellcolor{mitblue}\ding{51} &  \cellcolor{mitblue}\ding{51} &  \cellcolor{mitblue}\ding{51} 
		&  \cellcolor{mitblue}\textbf{1.25 }&  \cellcolor{mitblue}\textbf{0.70} 
		&  \cellcolor{mitblue}\textbf{4.85} &  \cellcolor{mitblue}\textbf{10.50} \\
		
		\bottomrule
	\end{tabularx}
\end{table}

\section{Conclusion}
In this paper, we address object removal in dense scenes, where instance interference severely degrades performance. 
We propose \textbf{DORS}, a training-free and plug-and-play framework that formulates object removal as semantic information flow control in the attention space. 
By integrating Instance-Filtered Attention and Context-Guided Routing, \textbf{DORS} effectively suppresses misleading information from similar instances while preserving structural consistency. 
We further introduce DOR-Bench, a benchmark for systematic evaluation in dense scenes.
Extensive experiments on DOR-Bench and public benchmarks demonstrate strong performance in removal accuracy, visual consistency, and generalization.

\begin{acks}
This research is supported by Institute of Advanced Medicine and Frontier
Technology (2023IHM01080), and sponsored by CCF-NetEase ThunderFire
Innovation Research Funding (NO. CCF-Netease 202513); The computation
is completed on the HPC Platform of Hefei University of Technology.
\end{acks}


\balance
\bibliographystyle{ACM-Reference-Format}
\bibliography{sample-base}

\clearpage
\appendix
\section*{Appendix}
This supplementary material provides detailed implementation, extended experimental results, and additional analyses to further support the effectiveness and robustness of the proposed \textbf{DORS} framework. The contents are organized as follows:
\begin{itemize}
	\item Full pipeline description, presenting the complete inference procedure and attention control mechanism of \textbf{DORS} (Sec.~\ref{app:pipeline});
	
	\item Dataset construction of DOR-Bench, detailing data sources, annotation strategy, and the characteristics of dense scenes with strong instance interference (Sec.~\ref{app:dor});
	
	\item Additional qualitative results, providing extended visual comparisons on DOR-Bench and RORD to illustrate typical failure modes and improvements (Sec.~\ref{app:more_results});
	
	\item User study details, including the human evaluation protocol, scoring criteria, and VLM-based automated assessment (Sec.~\ref{app:user_study});
	
	\item Efficiency comparison, reporting inference latency and GPU memory usage under consistent settings (Sec.~\ref{app:efficiency});
	
	\item Additional ablation results, offering further analysis of component contributions and their interactions in dense scenes (Sec.~\ref{app:ablation});
	
	\item Design ablation, offering further analysis of component contributions, fusion strategies, and routing designs (Sec.~\ref{app:designablation});
	
	\item Parameter analysis, investigating the effect of the density control parameter and its impact on performance (Sec.~\ref{app:parameter});
	
	\item Effect of segmentation models, evaluating the robustness of \textbf{DORS} under different segmentation models and oracle masks (Sec.~\ref{app:segmentation});
	
	\item Related work, providing additional discussions on diffusion-based object removal, attention control, and training-free image editing methods (Sec.~\ref{app:related});
	
	\item Failure cases, discussing challenging scenarios and limitations of \textbf{DORS} (Sec.~\ref{app:failure});
	
\end{itemize}

\begin{algorithm}[!b]
	\caption{Inference Pipeline of \textbf{DORS}}
	\label{alg:dors}
	\small
	\KwIn{input image $I$, target mask $M$, diffusion model $\epsilon_\theta$, text condition $c=\varnothing$, total steps $T$, encoder $E$, decoder $D$, segmentation model}
	
	\textbf{Initialization} \\
	Construct masked image $I_m = I \odot (1 - M)$\;
	
	Extract similar-instance regions $M_s$ using segmentation model\;
	
	Resize masks to obtain $M^{(r)}, M_s^{(r)}$ (for attention control) and $M^{(l)}$ (for diffusion conditioning)\;
	
	Encode into latent space:
	$\bar{z}_0 = E(I),\; z_m = E(I_m)$\;
	
	Initialize latent variable:
	$z_T \sim \mathcal{N}(0, \mathbf{I})$\;
	
	\For{$t = T$ \KwTo $1$}{
		
		\If{$t > T/2$}{

			Construct  full-context bias $B_f$ and pruned bias $B_p$ according to Eq.~(\ref{eq:B_f}) and Eq.~(\ref{eq:B_p})\
			
			Compute attention outputs from the  full-context and pruned pathways,  $O_f$ and $O_p$, via Eq.~(\ref{eq:O_f_}) and Eq.~(\ref{eq:O_p})\;
			
			Compute dynamic weight $w(x)$ via Eq.~(\ref{eq:rho})--(\ref{eq:w_p})\;

			Obtain fused attention $O$ via Eq.~(\ref{eq:O_final})
			
			Replace the standard self-attention computation in $\epsilon_\theta$ with the fused attention, yielding a modified predictor $\epsilon_\theta'$;
			
			Set active predictor:
			$\bar{\epsilon}_\theta = \epsilon_\theta'$\;
		}
		\Else{
			Set active predictor:
			$\bar{\epsilon}_\theta = \epsilon_\theta$\;
		}
		
		Predict noise:
		$\epsilon_t = \bar{\epsilon}_\theta([z_t, z_m, M^{(l)}], t, c)$\;
		
		Update latent:
		$z_{t-1} = \mathrm{Step}(z_t, \epsilon_t, t)$\;
		
		Obtain background latent:
		$\bar{z}_{t-1}$ via forward diffusion on $\bar{z}_0$\;
		
		Apply latent blending:
		$z_{t-1} = M^{(l)} \odot z_{t-1} + (1 - M^{(l)}) \odot \bar{z}_{t-1}$\;
	}
	
	Decode final latent:
	$I' = D(z_0)$\;
	\KwOut{object-removed image $I'$}
\end{algorithm}

\section{Full Pipeline Description}
\label{app:pipeline}
We present the complete inference pipeline of \textbf{DORS} in Algorithm~\ref{alg:dors}, which explicitly formulates object removal as controlled information propagation in the attention space. 
The framework integrates instance-aware attention suppression and Context-Guided Routing within the diffusion process, providing a structured and reproducible view of how removal is performed in dense scenes.
Operating at inference time, \textbf{DORS} requires no additional training and can be directly applied as a plug-and-play module. 
At each denoising step, \textbf{DORS} selectively regulates attention propagation in regions affected by instance interference. 
Instance-Filtered Attention suppresses misleading interactions with similar instances, effectively neutralizing ambiguous semantic signals, 
while Context-Guided Routing selectively restores relevant spatial context to preserve structural continuity. 

Together, these two components form a complementary mechanism: IFA removes erroneous semantic influence, while CGR reintroduces useful structural information. 
This design enables fine-grained control over attention information flow, resulting in stable and visually consistent removal results.

\section{Details of DOR-Bench}
\label{app:dor}

The proposed Dense Object Removal Benchmark (DOR-Bench) consists of 400 image-mask pairs, sourced from publicly available datasets~\cite{rorem, smarteraser}. 
While existing benchmarks mainly focus on generic object removal scenarios, they are not designed to capture the challenges posed by dense scenes with severe instance interference. 
As a result, they often fail to adequately evaluate methods under such conditions.

To address this limitation, DOR-Bench is specifically constructed to target scenarios with strong semantic ambiguity caused by similar instances. 
We strategically select samples with high instance density and significant intra-class similarity, and re-curate the removal targets to emphasize cases where the target object is surrounded by similar distractors. 
The detailed statistics of DOR-Bench, including object categories, scene types, mask ratios, and instance density distributions, are summarized in Table~\ref{tab:dorbench_stats}.
This design explicitly stresses the failure mode of erroneous information propagation in attention, which is central to the problem addressed in this work.

For annotation, all target masks are re-annotated to ensure high accuracy and consistency. 
Initial masks are generated using state-of-the-art segmentation tools, followed by strict pixel-level manual refinement to improve boundary alignment and avoid background leakage. 
For objects with complex structures or heavy occlusions, multiple rounds of manual verification and cross-checking are conducted to further ensure annotation quality.
Overall, DOR-Bench provides a challenging and targeted benchmark for evaluating object removal in dense scenes, where controlling attention propagation is essential. 
Additional qualitative examples are provided in Fig.~\ref{fig6:More-Qualitative}, offering a more comprehensive visualization of the dataset and its inherent challenges.

\begin{table}[b]
	\centering
	\caption{
		Statistics of the proposed DOR-Bench.
		The benchmark is summarized from multiple perspectives, including object categories, scene types, mask ratio distributions, and instance density levels, highlighting the diversity and challenges of dense object removal scenarios.
	}
	\label{tab:dorbench_stats}
	\resizebox{1.0\linewidth}{!}{
		\begin{tabular}{l|lcc || l|lcc}
			\toprule
			\textbf{Aspect} & \textbf{Group} & \textbf{\#} & \textbf{Ratio} & \textbf{Aspect} & \textbf{Group} & \textbf{\#} & \textbf{Ratio} \\
			\midrule
			\multirow{5}{*}{Category}
			& Food  & 219 & 54.75\% & \multirow{5}{*}{Mask ratio}
			& $<1\%$ & 54 & 13.50\% \\
			& Animal & 62 & 15.50\% & & $1$--$3\%$ & 102 & 25.50\% \\
			& Book & 30 & 7.50\% & & $3$--$5\%$ & 118 & 29.50\% \\
			& Plant & 27 & 6.75\% & & $5$--$15\%$ & 114 & 28.50\% \\
			& Object & 62 & 15.50\% & &$>15\%$ &12 &3.00\% \\
			\midrule
			\multirow{3}{*}{Scene}
			& Indoor & 239 & 60.00\% & \multirow{3}{*}{Sim. density}
			& Low ($1$--$6$) & 67 & 16.75\% \\
			& Store & 52 & 12.75\% & & Medium ($7$--$10$) & 78 & 19.50\% \\
			& Natural & 109 & 27.25\% & & High ($>10$) & 255 & 63.75\% \\
			\bottomrule
	\end{tabular}}
\end{table}

\section{Additional Qualitative Results}
\label{app:more_results}

\begin{figure*}[htbp]
	\centering
	\includegraphics[width=\linewidth]{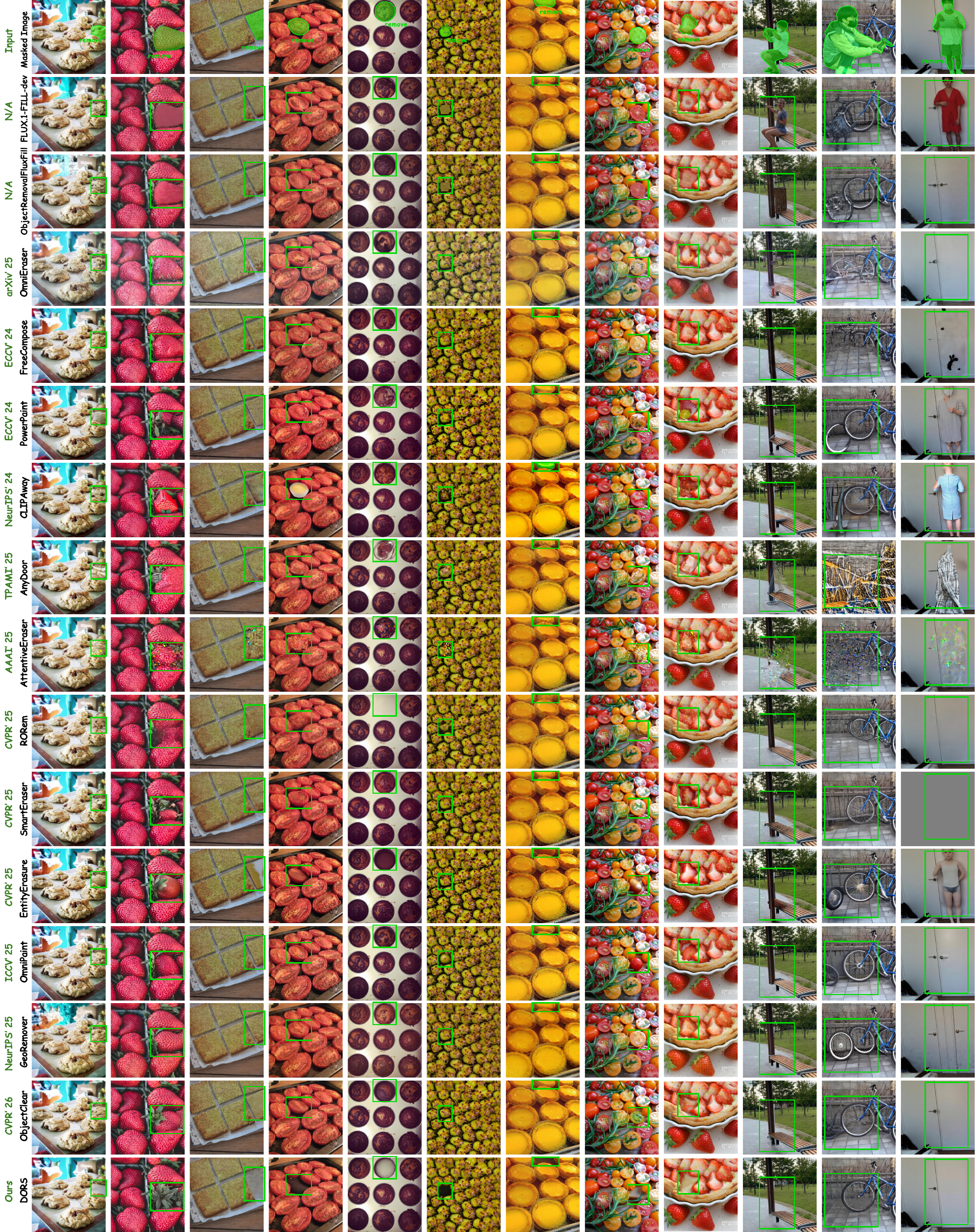}
	\caption{
		Additional qualitative results under various challenging scenarios. 
		Zoom in for better visualization of fine details.
	}
	\label{fig6:More-Qualitative}
\end{figure*}

We present additional qualitative results on DOR-Bench and RORD in Fig.~\ref{fig6:More-Qualitative}, 
further demonstrating the effectiveness of the proposed method under challenging conditions.
Existing methods exhibit several typical failure modes, including incomplete removal, hallucinated artifacts, over-smoothing, and background inconsistency. 
These failures are largely caused by uncontrolled attention propagation, where masked queries incorrectly attend to similar instances, leading to erroneous semantic transfer.
In contrast, our method explicitly mitigates such issues by regulating attention flow. 
It effectively suppresses misleading interactions from similar instances while preserving relevant structural and contextual information. 
As a result, the proposed method produces cleaner and more stable outputs, with fewer artifacts and better preservation of both structure and texture.
Even in highly challenging scenarios with densely distributed similar objects, our method maintains consistent visual quality and avoids common failure cases, demonstrating strong robustness and generalization ability.

\section{Details of User Study}
\label{app:user_study}

We provide a comprehensive description of the human perceptual evaluation protocol and the automated vision-language model (VLM) assessment used to evaluate object removal performance.

\begin{figure}[tbp]
	\centering
	\includegraphics[width=\linewidth]{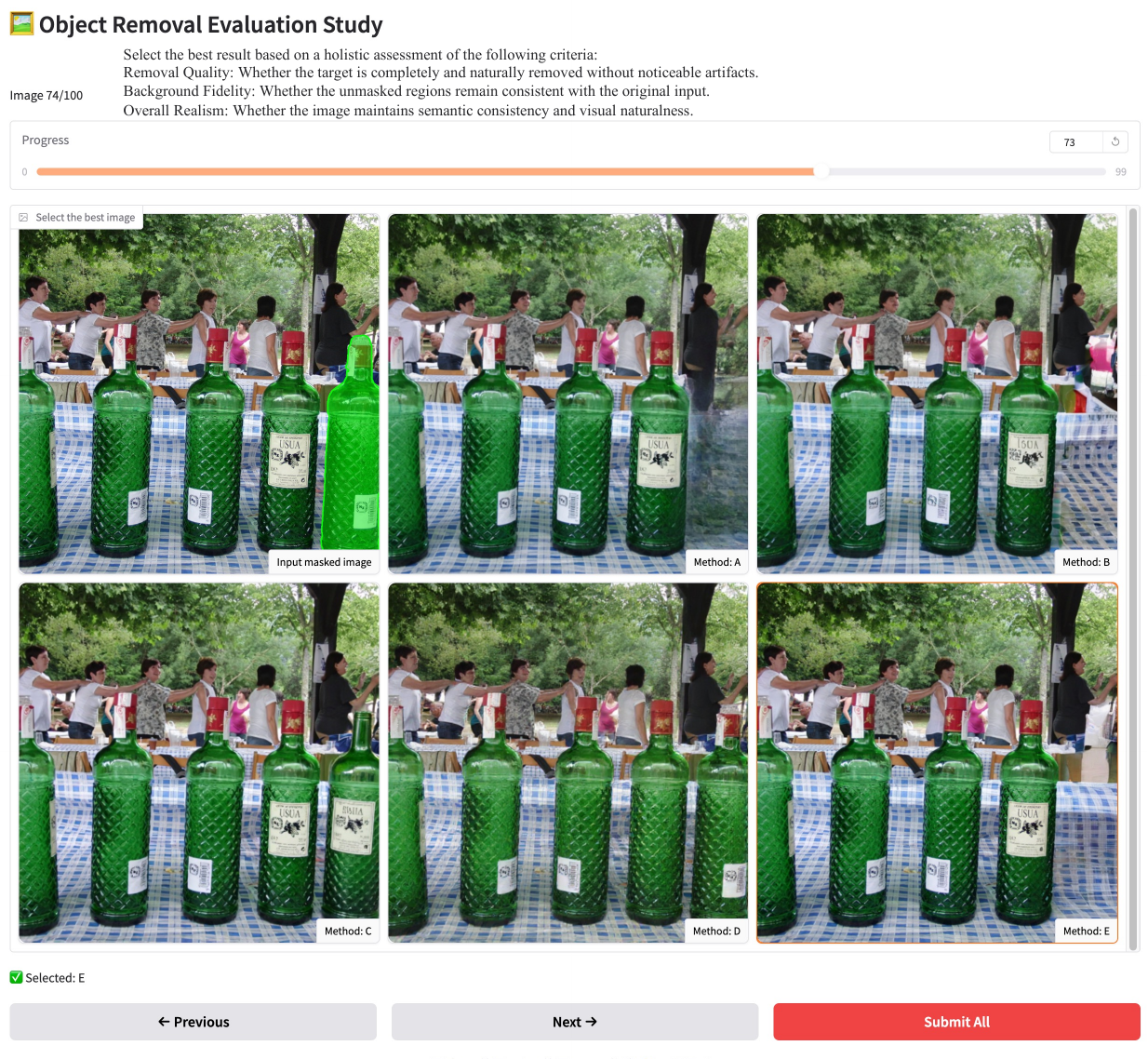}
	\caption{
		User study interface. The input image with the target mask is shown as a reference, 
		while candidate results are presented in randomized order with anonymized labels. 
	}
	\label{fig7:user_interface}
\end{figure}

\noindent
\textbf{Human Evaluation Protocol.}
For each evaluation sample, participants were presented with the \textit{input with mask} (the original image with the target region highlighted in green) to specify the removal area. Alongside this reference, the outputs of five methods were displayed, including our method and four representative baselines. To reduce subjective bias, all candidate results were anonymized and presented in randomized order, while the reference image remained fixed as a visual anchor.
We recruited 10 non-expert participants, each completing 100 independent evaluation tasks. For each task, participants were instructed to select the best result based on a holistic assessment of the following criteria:

{
	\leftmargini=4mm 
	\begin{itemize}
		\item \textbf{Removal Quality:} Whether the target is completely and naturally removed without introducing noticeable artifacts.
		\item \textbf{Background Fidelity:} Whether the unmasked regions remain consistent with the original input.
		\item \textbf{Overall Realism:} Whether the image preserves semantic consistency and visual naturalness.
	\end{itemize}
}
If all candidates failed to meet basic quality standards, participants were allowed to leave the sample unselected. For each method, we computed the selection frequency per participant and then averaged across participants to obtain the final \textbf{Mean Preference Rate}.
To enhance transparency, a screenshot of the evaluation interface is shown in Fig.~\ref{fig7:user_interface}. The input with mask is displayed as a fixed reference, while candidate results are presented with anonymized labels (e.g., A, B, C) in randomized order. Participants selected the best result via clicking, or left the sample unselected if all outputs were unsatisfactory. 

\noindent
\textbf{VLM-based Automated Evaluation.}
To provide a scalable and reproducible evaluation signal, we further employ the vision-language model Qwen3.5-Plus~\cite{qwen3.5} as an automated perceptual evaluator.
Qwen3.5-Plus is a large-scale multimodal model (approximately 397B parameters) with strong visual reasoning capabilities.

The model is evaluated under the same input format and criteria as the human study.
For each sample, it receives the masked input and all candidate results, and follows a unified evaluation prompt considering object removal completeness, background preservation, visual naturalness, and semantic consistency.
The detailed prompt template is provided in Table~\ref{tab:vlm_prompt}.
To mitigate the inherent position bias of VLMs, the order of candidates is randomly shuffled across five independent evaluation rounds.

For each candidate, the model outputs a score from 1 to 10 together with a brief explanation.
A method is selected if it achieves a clearly higher score than other candidates; otherwise, when all candidates receive low scores (e.g., below 5), no selection is made, following the human evaluation protocol.
The final results are aggregated through majority voting.

Furthermore, we analyze the consistency between VLM-based evaluation and human judgment.
Both human and VLM evaluations rank DORS as the Top-1 method in Table~4 of the main paper, achieving a Spearman rank correlation coefficient of 0.90 over five methods.
Since a value closer to 1 indicates stronger agreement between rankings, this result demonstrates the high consistency between VLM-based evaluation and human preference.

\begin{table}[t]
	\centering
	\setlength{\abovecaptionskip}{2pt}
	\setlength{\belowcaptionskip}{2pt}
	\caption{
		Prompt template used for the VLM-based automated evaluation.
		The prompt guides Qwen3.5-Plus to compare candidate results according to predefined criteria, including object removal completeness, background preservation, visual naturalness, and semantic consistency.
	}
	\label{tab:vlm_prompt}
	\small
	\begin{tabular}{p{0.92\linewidth}}
		\toprule
		
		\textbf{Role.}
		You are a professional image quality assessment expert.
		Please evaluate the object removal performance of the given images.

		\textbf{Evaluation Criteria.}
		
		\begin{itemize}
			\item Object Removal Completeness
			\item Background Preservation
			\item Visual Naturalness
			\item Semantic Consistency
		\end{itemize}

		\textbf{Image Description.}
		
		\begin{itemize}
			\item Image 1: Masked Image
			\item Image 2: Method A's result
			\item Image 3: Method B's result
			\item Image 4: Method C's result
		\end{itemize}
		
		\textbf{Task.}
		Carefully compare all candidate images according to the above criteria and:
		
		\begin{itemize}
			\item Select the best-performing method among all candidates.
			\item Provide a concise explanation for the selection.
			\item Assign a score from 1 to 10 for each method, where 10 indicates the best performance.
		\end{itemize}

		\textbf{Output Format.}
		
		\texttt{\{"best\_method": "<method\_name>",}\\
		\texttt{"reasoning": "<brief explanation>",}\\
		\texttt{"scores": \{"Method A": 8, "Method B": 6, "Method C": 4\}\}}
		
		\\
		\bottomrule
	\end{tabular}
\end{table}

\section{Full Efficiency Comparison}
\label{app:efficiency}

We present an efficiency comparison of all methods in terms of inference latency and GPU memory footprint. 
All methods are evaluated on the same hardware using their official implementations with default configurations. The input resolution is kept consistent across all methods for fair comparison.
Inference latency is reported as the average per-image processing time (seconds per image), and GPU memory corresponds to the peak VRAM usage during inference. 
As shown in Table~\ref{tab10:resource_consumption}, our method achieves a favorable balance between efficiency and resource consumption without additional training, demonstrating competitive performance among both training-based and training-free approaches.

\section{Additional Ablation Results}
\label{app:ablation}

\begin{figure}[htbp]
	\centering

	\includegraphics[width=\linewidth]{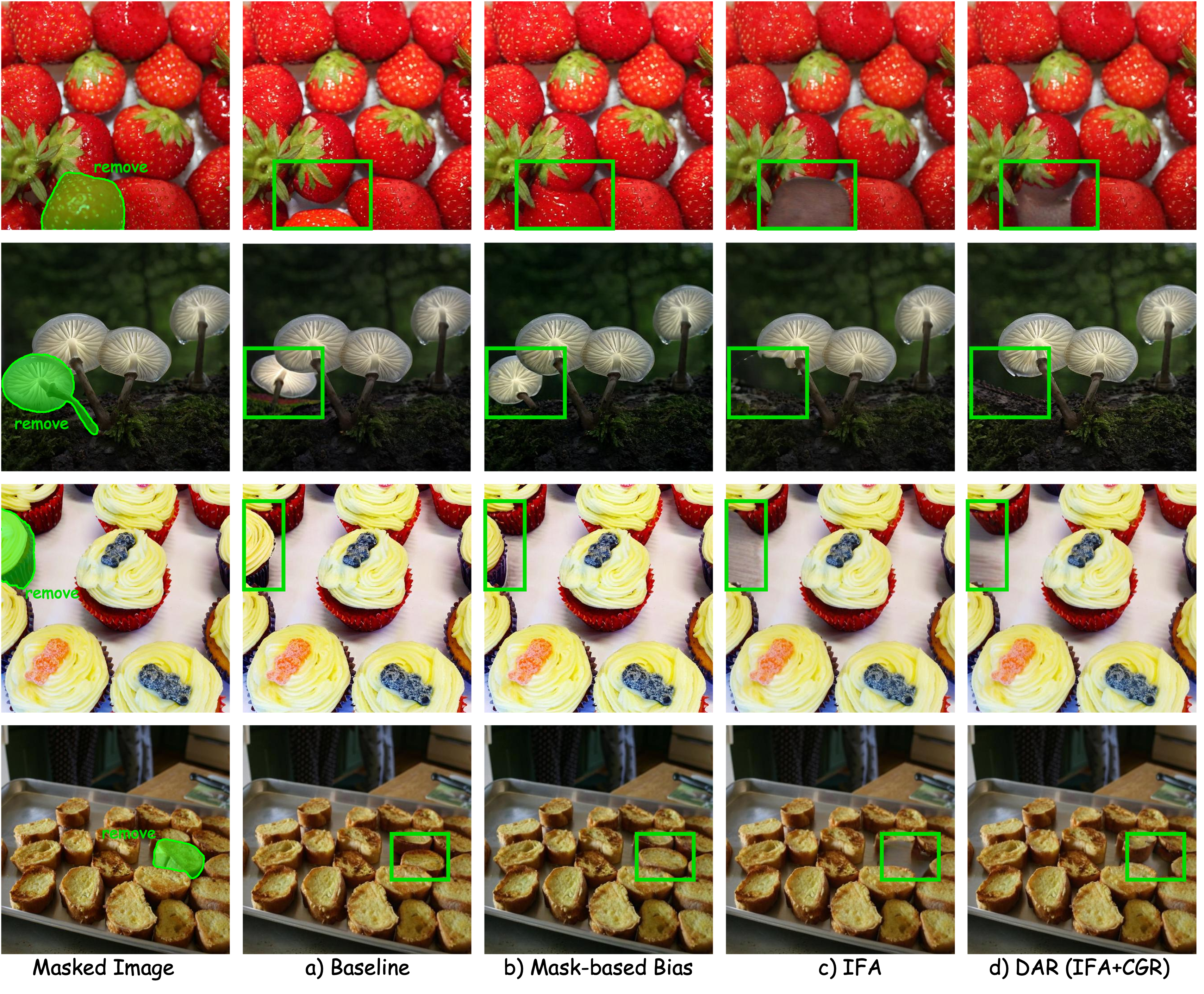}

	\caption{
		Additional qualitative ablation results. 
		Mask-based Bias suppresses intra-mask propagation but still suffers from interference from similar instances, while IFA mitigates such interference at the cost of structural degradation. 
		By integrating IFA and CGR, DAR effectively suppresses interference from similar instances while restoring valid contextual information, improving overall consistency.
	}
	\label{fig6: app-ab}

\end{figure}

We present additional qualitative ablation results in Fig.~\ref{fig6: app-ab} to further analyze the role of each component and their interactions in dense scenes. 
The comparison is conducted in a progressive manner, including the baseline, the full-context branch (FCB), the proposed Instance-filtered Attention (IFA), and the proposed Context-Guided Routing (CGR), consistent with the component settings reported in Table~\ref{tab3:ablation_components}.
From the results, we observe that the baseline suffers from severe instance interference in dense scenes, due to the lack of explicit control over attention propagation, leading to incomplete removal or incorrect reconstruction. 
With the introduction of FCB, intra-mask propagation is effectively suppressed. 
However, since it still relies on global context aggregation, similar instances in unmasked regions can still influence the target region through attention, resulting in removal failure or unstable generation.
Building upon this, IFA performs instance-level filtering on attention connections, which effectively blocks misleading interactions from similar instances and produces cleaner removal results. 
However, this strong constraint may also lead to over-pruning, where useful contextual information for structure recovery is removed, causing structural degradation or local inconsistency.
To address this issue, CGR introduces a spatially adaptive routing mechanism on top of IFA, which selectively restores suppressed information. 
This design preserves the ability to suppress interference while recovering necessary structural and contextual cues, resulting in more stable and consistent outputs. 
In particular, in challenging cases with multiple similar instances, CGR achieves a better balance between interference suppression and structural continuity.
Overall, these results demonstrate that IFA and CGR play complementary roles: IFA suppresses misleading semantic information from similar instances, while CGR restores useful structural cues. 
Their combination enables fine-grained control of attention information.

\begin{table}[htbp]
	\centering
	\caption{Efficiency comparison with State-of-the-Arts. Inference time (seconds per image) and GPU memory usage (GB) are reported. The best value for each metric is highlighted in \textcolor{red}{\textbf{red bold}}, and the second best is shown as \textcolor{blue}{\textbf{blue bold}}. }
	\small
	\renewcommand{\arraystretch}{1.2}
	\setlength{\tabcolsep}{4pt}
	\label{tab10:resource_consumption}
	\begin{tabularx}{\columnwidth}{l c c c}
		\toprule
		Method & Type & Time(s)$\downarrow$ & GPU(GB)$\downarrow$ \\
		\midrule
		ObjectRemovalFluxFill~\cite{object-removal-lora}       & Training-Based & 9.48 & 32.13 \\
		OmniEraser (arXiv'25)~\cite{omnieraser}       & Training-Based & 5.06 & 32.22 \\
		AnyDoor (TPAMI'25)~\cite{anydoor}         & Training-Based & 6.57 & 10.23 \\
		EntityErasure (CVPR'25)~\cite{entityerasure}    & Training-Based & 3.59 & 12.54 \\
		RORem (CVPR'25)~\cite{rorem}          & Training-Based & 3.72 & 8.96 \\
		OmniPaint (ICCV'25)~\cite{omnipaint}       & Training-Based & 9.56 & 32.07 \\
		GeoRemover (NeurIPS'25)~\cite{georemover}    & Training-Based & 33.87 & 72.35 \\
		ObjectClear (CVPR'26)~\cite{objectclear}      & Training-Based & \textcolor{red}{\textbf{1.69}}  &  \textcolor{blue}{\textbf{7.77}}  \\ 
		\midrule
		FLUX.1-Fill-dev~\cite{flux2024}                 & Training-Free & 8.79 & 32.05 \\
		FreeCompose (ECCV'24)~\cite{freecompose}     & Training-Free & 24.26 & 9.71 \\
		AttentiveEraser (AAAI'25)~\cite{attentiveeraser}  & Training-Free & \textcolor{blue}{\textbf{3.43}}  &  \textcolor{red}{\textbf{7.66}} \\
		\cellcolor{mitblue}\textbf{DORS (Ours)}     &  \cellcolor{mitblue}Training-Free &  \cellcolor{mitblue}4.85 &  \cellcolor{mitblue}10.50 \\
		\bottomrule
	\end{tabularx}
\end{table}

\section{Design Ablation}
\label{app:designablation}
We conduct a comprehensive ablation study to analyze the key design choices of the proposed Dynamic Attention Routing (DAR), including the routing strategy, weighting function, and attention control schedule.
As shown in Tab.~\ref{tab:design_ablation}, different routing strategies are first compared.
Equal weighting (row 1), which assigns fixed weights to the full-context and filtered pathways, achieves limited performance.
Introducing local spatial modeling through Distance-only routing (row 3) improves the MSN score from 6.00 to 4.75, demonstrating the importance of geometric relationships between the target region and similar instances.
Further incorporating global instance density with local distance in Dynamic routing (row 9) significantly improves performance (MSN 4.75 $\rightarrow$ 1.25), indicating that both global semantic interference and local spatial constraints are necessary for effective routing.
In contrast, Density-only routing (row 2) performs poorly (MSN 11.75), showing that global information without spatial awareness is insufficient.

We further evaluate different weighting functions under the Dynamic routing strategy.
The proposed Piecewise weighting (row 9, Eq.~(\ref{eq:D_i})) consistently outperforms Hard Cutoff (row 4) and Linear Decay (row 5).
This verifies that gradually adjusting the filtering strength according to spatial distance provides a better balance between semantic suppression and structural preservation.

Finally, we investigate the effect of attention control schedules.
Applying DAR in early denoising stages (2/4 steps, row 9) achieves the best performance, while applying it for shorter or longer ranges leads to degradation.
This observation is consistent with diffusion dynamics, where early denoising steps primarily determine global structures.
These results validate the effectiveness of each component in DAR.

\begin{table}[t]
	\centering
	\caption{
		Ablation study of the proposed dynamic fusion strategy, including different fusion strategies, weighting functions, and intervention schedules.
	}
	\label{tab:design_ablation}
	\small
	\setlength{\tabcolsep}{5pt}
	\renewcommand{\arraystretch}{1.1}
	\begin{tabular}{l l c c c}
		\toprule
		Fusion Strategy & Weighting & Schedule & MSN $\downarrow$ & MARS $\downarrow$ \\
		\midrule
		
		Equal         & Constant (0.5)         & 2/4 steps & 6.00  & 2.31 \\
		Density-only  & Constant ($w_{\max}$)  & 2/4 steps & 11.75 & 7.01 \\
		Distance-only & Piecewise              & 2/4 steps & 4.75  & 2.05 \\
		
		\midrule
		
		Dynamic & Hard Cutoff   & 2/4 steps & 5.25  & 2.17 \\
		Dynamic & Linear Decay  & 2/4 steps & 4.00  & 1.64 \\
		
		\midrule
		
		Dynamic & Piecewise & 1/4 steps & 23.75 & 11.69 \\
		Dynamic & Piecewise & 3/4 steps & 13.25 & 4.86 \\
		Dynamic & Piecewise & 4/4 steps & 20.50 & 10.60 \\
		
		\midrule
		\cellcolor{mitblue}\textbf{Dynamic} &  \cellcolor{mitblue}\textbf{Piecewise} &  \cellcolor{mitblue}\textbf{2/4 steps} &  \cellcolor{mitblue}\textbf{1.25} &  \cellcolor{mitblue}\textbf{0.70 }\\
		\bottomrule
	\end{tabular}

\end{table}

\section{Parameter Analysis}
\label{app:parameter}

We analyze the effect of the density control parameter $\alpha$ in Table~\ref{tab6:alpha}, which regulates the strength of suppression applied to similar-instance regions in the proposed routing mechanism.
When $\alpha$ is small, the suppression is weak, allowing attention to leak toward similar instances. 
This leads to residual artifacts and incomplete removal, as reflected by higher MSN and MARS values. 
As $\alpha$ increases, the model progressively suppresses such interference, resulting in improved removal quality and lower metric values. 
However, when $\alpha$ becomes too large, the suppression becomes overly aggressive, removing not only misleading semantics but also useful structural cues, which degrades reconstruction quality.
The best performance is achieved at $\alpha = 0.5$, where the model achieves a good balance between interference suppression and structural preservation. 
Overall, these results indicate that $\alpha$ effectively controls the trade-off between suppressing misleading information from similar instances and preserving useful structural context, which is essential for the proposed Context-Guided Routing mechanism.

\begin{table}[htbp]
	\centering
	\caption{Sensitivity analysis of the density control parameter $\alpha$ in the proposed Context-Guided Routing.  }
	\label{tab6:alpha}
	\small
	\setlength{\tabcolsep}{6pt}
	\renewcommand{\arraystretch}{1.15}
	\begin{tabularx}{0.6\columnwidth}{
			>{\centering\arraybackslash}p{1.2cm}
			| *{2}{>{\centering\arraybackslash}X}
		}
		
		\toprule
		$\alpha$
		& \shortstack{MSN $\downarrow$}
		& \shortstack{MARS $\downarrow$} \\
		
		\midrule
		
		0.1 & 6.75  & 3.02  \\
		0.2 & 5.50  & 2.56  \\
		0.3 & 7.25  & 2.73  \\
		0.4 & 4.50  & 2.19  \\
		
		\cellcolor{mitblue}0.5 &  \cellcolor{mitblue}\textbf{1.25} &  \cellcolor{mitblue}\textbf{0.70} \\
		0.6 & 4.75  & 2.47  \\
		0.7 & 4.00  & 1.89  \\
		0.8 & 3.50  & 1.79  \\
		0.9 & 4.75  & 2.85  \\
		1.0 & 5.25  & 3.35  \\
		
		\bottomrule
	\end{tabularx}
\end{table}

\section{Effect of Segmentation Models}
\label{app:segmentation}
Since \textbf{DORS} takes object masks as input, we investigate the impact of segmentation quality on removal performance.
Although our method is not tied to a specific segmentation model, mask quality may influence the accuracy of instance localization and attention routing.
We evaluate \textbf{DORS} using masks generated by different segmentation models, including SAM1, SAM2, SAM3, and manually annotated masks (Oracle).
The evaluation is conducted on a challenging subset containing dense scenes with multiple similar instances.
Mask quality is measured by Instance Recall, Instance Precision, and IoU, while removal performance is evaluated using MSN and MARS.
As shown in Table~\ref{tab:mask_quality}, higher-quality masks generally lead to better removal performance.
However, the performance gap between SAM3 and SAM1 remains limited, demonstrating the robustness of \textbf{DORS} to imperfect mask inputs.
Notably, even the weakest SAM1-based variant consistently outperforms existing state-of-the-art methods.

\begin{table}[t]
	\centering
	\caption{Effect of segmentation models on DORS.
		We compare masks generated by SAM1, SAM2, SAM3, and manually annotated masks (Oracle) on challenging dense-scene subset.}
	\label{tab:mask_quality}
	\resizebox{1.0\linewidth}{!}{
		\begin{tabular}{lccccc}
			\toprule
			\multirow{2}{*}{\textbf{Source}} 
			& \multicolumn{3}{c}{\textbf{Similar-instance mask quality}} 
			& \multicolumn{2}{c}{\textbf{Removal performance}} \\
			\cmidrule(lr){2-4} \cmidrule(lr){5-6}
			& \textbf{Instance Recall}$\uparrow$ 
			& \textbf{Instance Precision}$\uparrow$ 
			& \textbf{IoU}$\uparrow$ 
			& \textbf{MSN}$\downarrow$ 
			& \textbf{MARS}$\downarrow$ \\
			\midrule
			ObjectClear (CVPR'26)  & -  & -  & -  & 22.91 & 16.13 \\
			AttentiveEraser (AAAI'25) & -  & -  & -   & 18.75 & 6.68 \\
			\hline
			DORS (SAM1)      & 99.58 & 96.29 & 95.59  & 8.33 & 5.45 \\
			DORS (SAM2)      & 99.65 & 94.32 & 96.31  & 6.25 & 4.22 \\
			DORS (SAM3)      & 99.85 & 98.13 & 97.01  & 4.17 & 2.38 \\
			DORS (Oracle)    & 100.0 & 100.0 & 100.0  & 2.08 & 1.99 \\
			\bottomrule
	\end{tabular}}
\end{table}

\section{Related Work}
\label{app:related}
\noindent\textbf{Training-based Removal Methods.}
Most existing approaches follow a data-driven fine-tuning paradigm, adapting pre-trained models to object removal via paired supervision.
Prior works improve this paradigm along several directions, including enhanced data construction, stronger conditioning signals, and structural or geometric priors.
For instance, large-scale synthetic data construction and augmentation strategies improve data diversity~\cite{objectdrop, rorem, smarteraser, omnieraser, erasediff}, while prompt-based guidance and semantic representations enhance controllability~\cite{powerpaint, magiceraser, clipaway, entityerasure}.
Other works further introduce geometric priors~\cite{georemover, asuka} or unified editing frameworks to improve visual consistency~\cite{anydoor, omnipaint, paintbyinpaint}.

Despite these advances, such methods rely on learned input–output mappings and lack explicit control over cross-region information flow in the attention space.
Moreover, they are inherently constrained by the distribution of training data, which predominantly consists of sparse or isolated objects.
This limitation makes them difficult to generalize to dense scenes with multiple semantically similar instances, where cross-region interference becomes prominent.
As a result, features from visually similar regions may be incorrectly aggregated, leading to incomplete removal, structural inconsistency, and visible artifacts.

\noindent\textbf{Training-free Removal Methods.}
Training-free methods manipulate pre-trained models at inference time by modifying the generation process.
FreeCompose~\cite{freecompose} formulates object removal as a latent optimization problem, suppressing masked-region information by discarding key and value features in self-attention.
However, it leaves semantically similar instances in unmasked regions unaffected, allowing them to participate in global attention matching and causing erroneous reconstruction.
AttentiveEraser~\cite{attentiveeraser} instead modifies self-attention to encourage masked regions to attend to the background, combined with temperature scaling to smooth attention distributions.
While this reduces interference, it does not explicitly distinguish different semantic sources, which may introduce unnecessary suppression and compromise structural details.
These limitations indicate that effective object removal requires not only suppressing undesired information, but also selectively controlling cross-region information flow.

Different from these approaches, our method explicitly models information flow in the attention space and introduces dynamic attention routing to selectively regulate cross-region interactions, enabling precise object removal and consistent background completion in dense scenes.

\section{Failure Cases}
\label{app:failure}
\begin{figure}[htbp]
	\centering
	\includegraphics[width=0.9\linewidth]{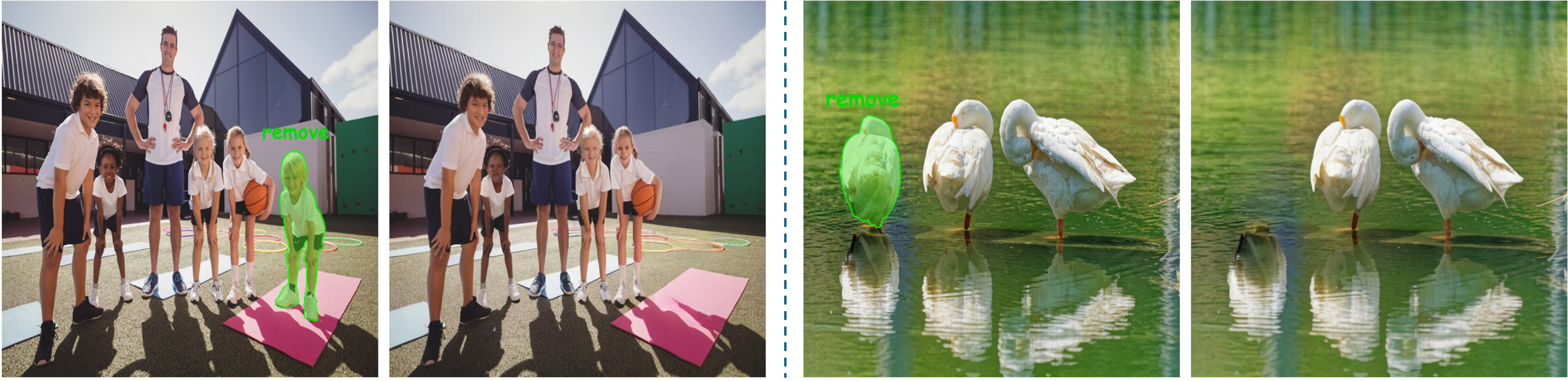}
	\caption{
		Failure cases of DORS on complex visual effects.
		DORS may fail to remove object-related effects, such as shadows and reflections.
		In (a), the shadow of the removed object remains after removal, while in (b), the reflection of the removed object is still visible.
	}
	\label{fig:failure}
\end{figure}

Although \textbf{DORS} effectively removes target objects from dense scenes with multiple similar instances and alleviates semantic interference from surrounding instances, it remains limited in handling complex object-associated visual effects beyond the target appearance itself. As shown in Fig.~\ref{fig:failure}, the target object can often be successfully removed, while related effects such as shadows, reflections, and local illumination changes may remain in the generated result. For example, a shadow cast by the removed object may persist on the background surface, reducing both visual realism and overall consistency. This limitation stems from the primary objective of \textbf{DORS}, which is to suppress misleading semantic information from similar instances while preserving useful contextual cues from unmasked regions for background reconstruction. It does not explicitly model physically coupled phenomena such as shadow formation, light transport, and reflection. Therefore, jointly identifying and removing target objects together with their associated visual effects remains an important direction for improving the completeness and realism of object removal in complex scenes.

\end{document}